\documentclass[journal]{IEEEtran}
\usepackage{amsmath,amsfonts}
\usepackage{algorithmic}
\usepackage{algorithm}
\usepackage{array}
\usepackage[caption=false,font=normalsize,labelfont=sf,textfont=sf]{subfig}
\usepackage{textcomp}
\usepackage{tabularx}
\usepackage{multirow} 
\usepackage{booktabs}
\usepackage{stfloats}
\usepackage{xcolor}
\usepackage{float}
\usepackage{url}
\usepackage{verbatim}
\usepackage{graphicx}
\usepackage{makecell} 
\usepackage{cite}
\usepackage[colorlinks=true, linkcolor=blue, citecolor=blue, urlcolor=blue]{hyperref}

\usepackage[table]{xcolor}
\definecolor{academicred}{RGB}{204, 51, 51} 

\newcommand{\gtext}[1]{\textcolor{green!60!black}{#1}}

\begin{document}

\title{GRASP: Guided Region-Aware Sparse Prompting for Adapting MLLMs to Remote Sensing}

\author{Qigan Sun, Chaoning Zhang,~\IEEEmembership{Senior Member, IEEE,} Jianwei Zhang, Xudong Wang, Jiehui Xie, Pengcheng Zheng, Haoyu Wang, Sungyoung Lee,~\IEEEmembership{Member, IEEE,} Chi-lok Andy Tai, Yang Yang,~\IEEEmembership{Senior Member, IEEE,} Heng Tao Shen,~\IEEEmembership{Fellow, IEEE}

\thanks{Qigan Sun, Xudong Wang, and Sungyoung Lee are with the School of Computing, Kyung Hee University, Yongin-si, South Korea (email: sunqigan0206@gmail.com;wl200203@khu.ac.kr; sylee@oslab.khu.ac.kr).}
\thanks{Chaoning Zhang, Jianwei Zhang, Jiehui Xie, Pengcheng Zheng and Yang Yang are with the School of Computer Science and Engineering, University of Electronic Science and Technology of China, Chengdu, China (email: chaoningzhang1990@gmail.com; zjw5428c@gmail.com; jiehuixie@std.uestc.edu.cn; zpc777@std.uestc.edu.cn; yang.yang@uestc.edu.cn).}
\thanks{Haoyu Wang is with the College of Computer Science and Information Engineering, Harbin Normal University, Harbin, 150025, China (email: w18724284923@outlook.com)}
\thanks{Chi-lok Andy Tai is with College of Professional and Continuing Education, The Hong Kong Polytechnic University, Hong Kong, China (email: andy.tai@cpce-polyu.edu.hk).}
\thanks{Heng Tao Shen is with School of Computer Science and Technology, Tongji University, Shanghai, China (email: shenhengtao@hotmail.com).}
}

\markboth{Submitted to IEEE Transactions on Geoscience and Remote Sensing}%
{Q. Sun \MakeLowercase{\textit{et al.}}: GRASP: Guided Region-Aware Sparse Prompting...}

\maketitle

\begin{abstract}
In recent years, multimodal large language models (MLLMs) have shown strong potential in visual question answering tasks. However, effectively adapting them to remote sensing visual question answering (RSVQA) remains challenging. Unlike natural images, remote sensing imagery typically exhibits large scale variation, sparse target distributions, extensive background regions, and complex spatial semantics. These characteristics cause existing fine-tuning and global prompting methods to overfit irrelevant background patterns and overlook task-relevant local details. To address this issue, we propose Guided Region-Aware Sparse Prompting (GRASP), a parameter-efficient fine-tuning (PEFT) framework for remote sensing MLLMs. GRASP associates learnable soft prompts with spatial blocks derived from a frozen visual token grid and further employs a question-guided sparse fusion mechanism to dynamically aggregate block-level prompts into a compact global prompt. This design highlights question-relevant regional information and suppresses irrelevant background interference without modifying the pretrained vision or language backbones. Experimental results on multiple RSVQA benchmarks show that GRASP achieves stable and effective adaptation across different MLLM backbones, delivering competitive performance with high parameter efficiency and highlighting the value of explicitly modeling region-level spatial structure for remote sensing tasks.
\end{abstract}


\section{Introduction}
\IEEEPARstart{R}emote Sensing Visual Question Answering (RSVQA) is a key task that integrates visual perception with language reasoning, aiming to enable natural language interaction with RS imagery~\cite{lobry2020rsvqa,huang2025survey}. It provides intuitive and interpretable decision support for object recognition, scene understanding, and spatial relationship analysis, and holds significant value for practical applications such as disaster monitoring, urban planning, and environmental 
\begin{figure}[!t] 
  \centering
  \includegraphics[width=1.0\linewidth]{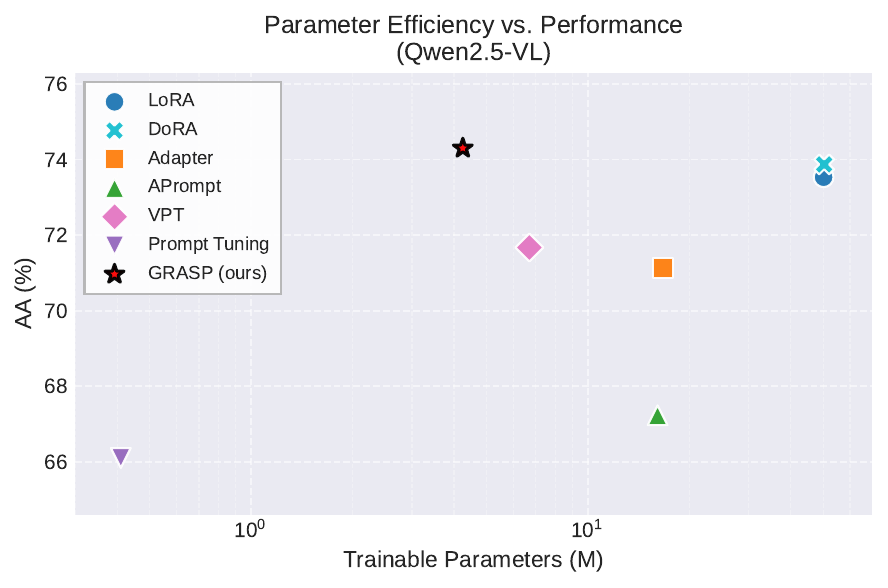}
  \caption{Parameter efficiency vs. performance on Qwen2.5-VL-7B. The y-axis denotes the overall accuracy (mean of the AA scores on RSVQA-LR, RSVQA-HR, and RSIVQA), plotted against trainable parameters (log scale). Our method, GRASP, is represented by a red star. Note that proximity to the top-left corner corresponds to higher accuracy with lower parameter costs.}
  \label{fig:qwen_performance}
\end{figure}
management~\cite{li2024vision,gomez2015multimodal, zhang2024jointly, zheng2023cgc}. Compared with visual question answering in natural scenes, RSVQA requires not only the recognition of basic visual elements such as object categories and spatial locations, but also an understanding of spatial context and semantic relationships within the image to perform high-level reasoning over complex questions~\cite{liu2024remoteclip,zhou2024towards}.

Traditional RSVQA methods typically adopt a pipeline consisting of a visual encoder and a task-specific module. For example, early works use convolutional neural networks (e.g., ResNet~\cite{he2016deep}) to extract image features and combine them with RNNs to process the text~\cite{lobry2020rsvqa}. Subsequent works introduce transformers to enhance the fusion and interaction of multimodal features~\cite{chappuis2022prompt,hackel2023lit}. These features are then passed through a classifier to generate the answers~\cite{li2024vision}. Although these methods achieve promising performance on specific datasets and task settings, they usually rely on predefined closed-form question sets, and their model architectures and output formats are highly task-dependent~\cite{songara2023visual,siripong2024large}. As a result, they lack scalability in open-domain scenarios, limiting their generality and extensibility.

In recent years, Multimodal Large Language Models (MLLMs) have significantly advanced cross-modal semantic alignment and language reasoning capabilities through large-scale multimodal pretraining, enabling a unified framework that supports diverse question formats and natural language-driven interaction~\cite{zhang2024earthmarker}. However, directly transferring general-purpose MLLMs to RS visual question answering tasks often leads to notable performance degradation~\cite{xiao2025foundation}. One primary reason is that general MLLMs are predominantly pretrained on large-scale natural image–text pairs and therefore lack prior knowledge of ground-object characteristics observed from a top-down perspective~\cite{li2025ddfav,xia2018dota}. This inherent training data distribution bias makes it difficult for such models to adapt to the substantial domain discrepancies between RS imagery and natural images, including profound object-scale variations, sparse object distributions, and complex region-level semantic structures. Furthermore, data distribution shifts from different sensors and spatiotemporal conditions further weaken model generalization, exacerbating the difficulty of aligning image content with question semantics~\cite{lobry2024visual,zhu2017deep}.

To adapt general-purpose MLLMs to remote sensing scenarios, recent studies mainly rely on domain-specific instruction tuning or generic parameter-efficient fine-tuning strategies~\cite{bazi2024rs,pang2024h2rsvlm,zhan2025skyeyegpt,hu2022lora}. These approaches demonstrate that transferring general multimodal backbones to the RS domain is feasible, but they still largely inherit adaptation mechanisms originally developed for natural-image settings~\cite{kuckreja2024geochat,zhang2024earthgpt}. 
In remote sensing imagery, task-relevant cues are often sparse and localized, while expansive background regions dominate the scene-level spatial layout. Under such conditions, spatially uniform adaptation may place insufficient emphasis on local details that are important for answering region-sensitive questions.

In this context, soft prompting offers a particularly promising direction because it adapts the model through a small number of learnable input vectors, introduces minimal trainable parameters, and preserves the pretrained vision and language backbones~\cite{lester2021power}. These properties make it an appealing starting point for parameter-efficient adaptation in remote sensing scenarios. 
However, conventional prompt-based methods, represented by VPT~\cite{jia2022visual} and APrompt~\cite{wang2023aprompt}, mostly adopt globally shared or layer-wise strategies and lack explicit modeling of region-level spatial structures. Consequently, these prompt parameters can only interact uniformly with all visual tokens in a sequential manner, making it difficult to establish structure-aware adaptation for different image regions. This design has clear limitations in remote sensing scenarios. Vision Transformer (ViT) encoders adopt fixed-granularity patch-level discretized representations. Although such representations preserve local visual information, their spatial partitioning scheme struggles to maintain the continuity of large-scale geographic entities, leading to the dispersion of complete local context across multiple independent tokens, which prevents prompt parameters from effectively capturing structurally consistent regional semantics. Meanwhile, core task objects in remote sensing images are typically highly sparse, while large homogeneous background regions dominate the scene. Without a region selection mechanism, weak target signals are easily overwhelmed by dominant background noise. In addition, different remote sensing visual question answering (RSVQA) tasks often rely on different local regions within the same image.
Therefore, there is an urgent need for a region-aware prompt tuning paradigm that can effectively integrate spatial structural priors with query semantic guidance, thereby enabling dynamic adaptation to task-relevant regions.

To this end, we propose Guided Region-Aware Sparse Prompting (GRASP) to bridge this gap by introducing an explicit spatial structural prior into soft prompt tuning. Specifically, by associating each spatial block derived from the visual token grid with a learnable soft prompt, our method instantiates these anchors to capture localized semantic cues in a structured and distributed manner. 
These block-level prompts are dynamically modulated according to question-conditioned regional relevance, and are subsequently aggregated into a compact global prompt through a question-guided sparse fusion mechanism. The resulting global prompt is injected as an additional visual token, providing structured visual guidance for the language model without modifying the pretrained vision or language backbones. This design enhances region-aware spatial understanding while maintaining architectural simplicity and strong parameter efficiency. We conduct extensive experiments on multiple RSVQA benchmarks, including ablation studies on different spatial block configurations. The results show that GRASP achieves competitive performance in answer accuracy and parameter efficiency compared to mainstream fine-tuning and prompt-based approaches, highlighting its effectiveness for RSVQA (see Fig.~\ref{fig:qwen_performance}). Our contributions are three-fold:
\begin{itemize}
    \item We propose Guided Region-Aware Sparse Prompting (GRASP), a PEFT strategy for RS MLLMs that associates learnable soft prompts with spatial blocks to enable region-aware spatial adaptation while keeping the vision and language backbones frozen.
    \item A question-guided sparse fusion mechanism is introduced to dynamically aggregate spatial context into a compact, task-specific global prompt. This design allows the model to selectively attend to question-relevant regions while effectively filtering out background noise.
    \item Extensive experiments on multiple RSVQA benchmarks demonstrate that GRASP achieves competitive performance among fine-tuning and prompt-based approaches, while maintaining high parameter efficiency.
\end{itemize}

\section{Related Work}
\subsection{RS Visual Question Answering}
RSVQA lies at the intersection of image understanding, spatial reasoning, and cross-modal semantic modeling. It aims to interpret natural language queries concerning object categories, spatial distributions, and semantic relationships within high-resolution RS imagery~\cite{chappuis2022prompt}. As an important decision-support tool for applications such as disaster monitoring and urban planning, RSVQA poses challenges that are substantially different from those encountered in natural scene VQA. The pronounced scale variation, sparse and uneven object distribution, and complex contextual semantics inherent to RS data significantly hinder effective cross-modal alignment and accurate answer generation~\cite{yuan2022exploring}.

Early studies in RSVQA primarily focus on dataset construction and benchmark establishment, alongside the adaptation of conventional visual question answering architectures. To facilitate supervised learning, datasets such as RSVQA~\cite{lobry2020rsvqa} and RSIVQA~\cite{zheng2021mutual} are constructed through automatic template-based question–answer generation. On the methodological side, initial approaches largely followed the paradigms of generic VQA systems, employing convolutional neural networks for visual feature extraction and recurrent models such as LSTMs for question encoding~\cite{antol2015vqa,xu2015show}. To better capture spatial dependencies and relational cues, subsequent works introduce region-level attention mechanisms~\cite{silva2022remote} and Transformer-based cross-modal architectures inspired by advances in vision–language modeling~\cite{lu2019vilbert}. Despite these improvements, these models represent a closed-set learning paradigm, often lacking the broad commonsense knowledge and robust reasoning capabilities required to fully interpret the intricate semantic relationships in diverse RS scenarios~\cite{feng2024multi}.

More recently, research attention has shifted toward adapting MLLMs to the RS domain~\cite{sun2024pixels,zhang2025earthgpt}. By leveraging large-scale vision–language foundation models and fine-tuning them on domain-specific datasets, recent studies report notable gains in cross-modal semantic alignment and reasoning performance~\cite{siebert2022multi, wang2024earthvqa}. Nevertheless, the practical deployment of these models is frequently impeded by substantial computational resources and storage overheads. The high cost of extensive model adaptation and prolonged training cycles reduces their flexibility for rapid or resource-constrained applications. As a result, achieving a balance between effective spatial semantic modeling and adaptation efficiency remains an open challenge. In this work, we address this issue by exploring a lightweight, spatially-aware design that captures local semantic information without incurring the overhead of full-parameter adaptation.

\subsection{Fine-tuning Strategies for MLLMs}
With the rapid development of multimodal learning, researchers have focused on fine-tuning large-scale multimodal models to integrate heterogeneous data, thereby improving their performance on specific tasks~\cite{zhang2024mm,yin2024survey}. Existing fine-tuning methods can be broadly categorized into full fine-tuning and PEFT~\cite{han2024parameter,ding2023parameter}.

\textbf{Full Fine-tuning.} Traditional full fine-tuning updates all parameters to enhance semantic alignment~\cite{wang2025parameter}. While effective for general instruction-tuning, this approach encounters a severe domain adaptation challenge in remote sensing~\cite{liu2024improved}. MLLMs are primarily pre-trained on ground-level natural images, while RS data exhibit a drastic perspective domain shift (bird's-eye view) and different physical characteristics~\cite{zhou2025advances,ou2025geopix}. Under limited RS supervision, full fine-tuning can be prone to overfitting and may compromise previously acquired general capabilities, while still facing difficulty in fully bridging the substantial domain gap between natural-image pre-training and remote sensing imagery~\cite{zhai2023investigating}.

\textbf{Parameter-Efficient Fine-tuning (PEFT).} To alleviate resource constraints, PEFT methods freeze the pretrained backbone and update only a small subset of parameters~\cite{houlsby2019parameter}. Representative approaches include Adapters~\cite{houlsby2019parameter}, LoRA, which approximates weight updates via low-rank factorization, and Prompt Tuning, which adapts models by optimizing learnable context representations~\cite{lester2021power}. In the RS domain, methods such as C-SAW~\cite{bhattacharya2023c} and MVP~\cite{zhu2024mvp} further leverage prompts to mitigate domain shift. 

However, a mismatch remains between these generic PEFT strategies and the characteristics of RS imagery~\cite{tian2025meta}. In remote sensing scenes, task-relevant information is often sparse and localized, while large background regions dominate the visual field~\cite{chen2025dynamicvis}. Under such conditions, globally shared adaptations, as in LoRA-like methods, may be insufficient to emphasize fine-grained local semantics, since task-relevant cues can be easily diluted by dominant background content~\cite{zidi2025lola}. Similarly, most prompt-based methods mainly perform static, image-level adaptation~\cite{xiao2025prompt,zhou2022conditional}. However, RSVQA requires dynamic spatial-semantic alignment, since answering a question often depends on identifying question-relevant regions rather than uniformly adapting the whole image representation~\cite{zhao2025text}. Moreover, RS data often vary across sensors, spatial resolutions, and acquisition conditions, which further weakens the transferability of generic adaptation strategies under limited supervision~\cite{bi2025ringmoe,xiao2025foundation}. These limitations motivate the need for a more spatially explicit adaptation mechanism, which is the focus of our proposed GRASP strategy.

\section{Methodology}
\begin{figure}[!t]
    \centering
    \includegraphics[width=1\linewidth]{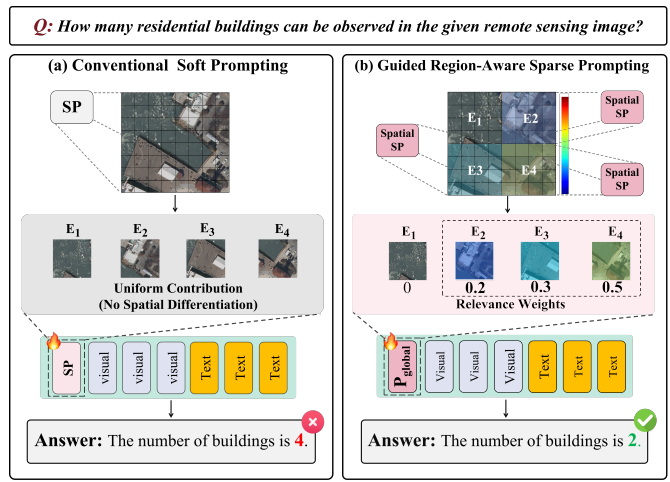}
    \caption{Conceptual comparison between Conventional soft prompting and GRASP. (a) Conventional soft prompting: Relies on static, uniform prompts. It struggles to suppress background interference (e.g., $E_1$), as it lacks region-aware adaptation. (b) GRASP: Utilizes question-guided sparse weighting to suppress irrelevant backgrounds ($E_1$ weight=0) and emphasize targets ($E_4$ weight=0.5). The resulting global prompt $P_{global}$ is computed by weighted fusion over all regions, where regions with zero weights are ignored.} 
    \label{fig:sp_concept}
\end{figure}
\begin{figure*}[ht!]
    \centering
    \includegraphics[width=1\linewidth,trim=0 0mm 0 0mm,clip]{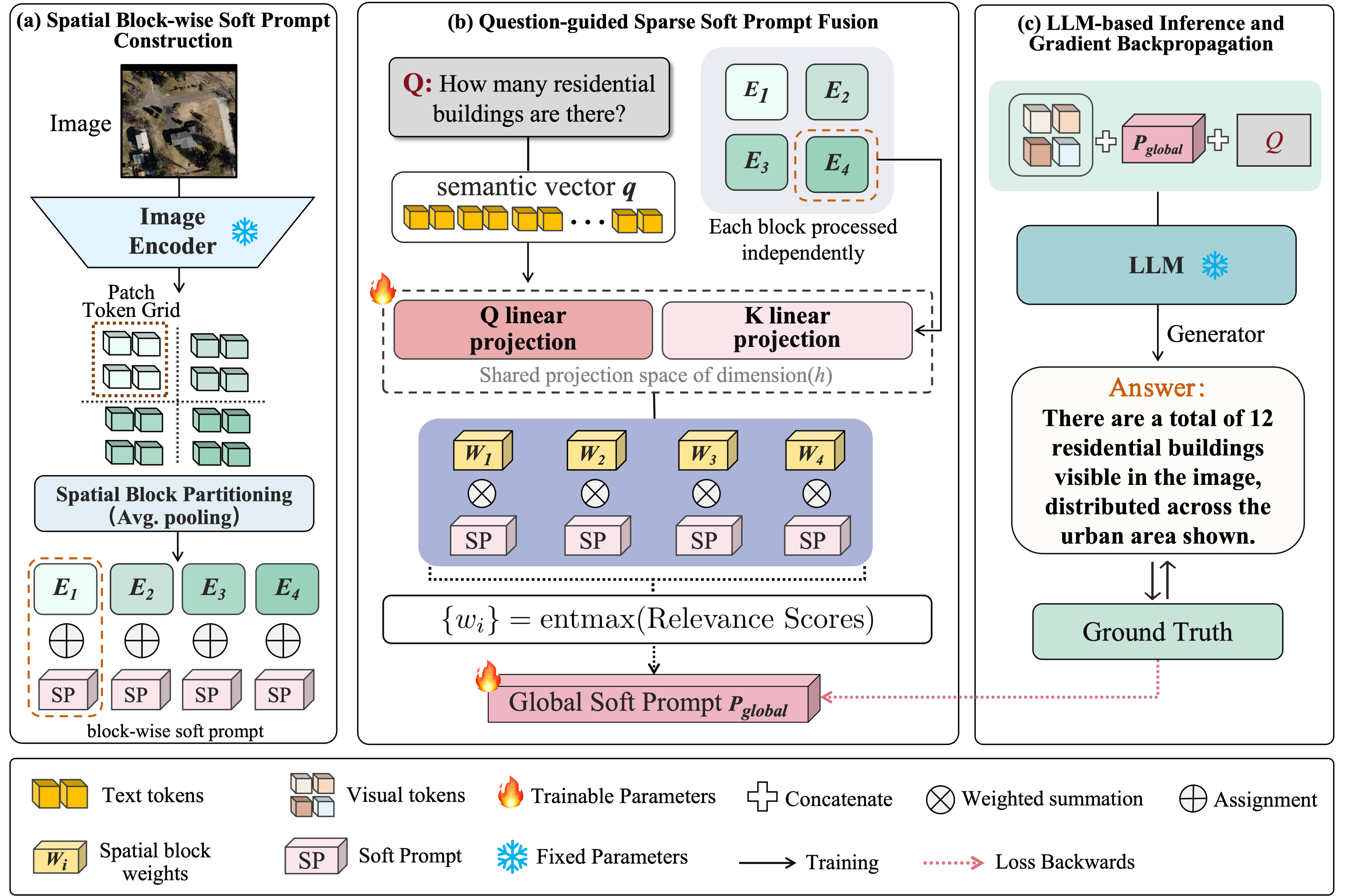}
    \caption{Architecture of the GRASP framework.
\textbf{(a) Spatial Block-wise Soft Prompt Construction:} The frozen vision encoder generates a visual token grid from the image. Average pooling partitions this grid into spatial blocks (e.g., $E_1$--$E_4$), where each block $E_i$ links to a learnable local soft prompt (SP).
\textbf{(b) Question-guided Sparse Soft Prompt Fusion:} The frozen LLM derives a semantic vector $q$ from the question. The system projects $q$ and block-level features $E_i$ into a shared space to compute relevance scores. Entmax activation converts these scores into sparse weights $\{w_i\}$, which the model uses to aggregate local SPs into a task-specific global soft prompt $P_{global}$.
\textbf{(c) LLM-based Inference and Gradient Backpropagation:} The framework injects $P_{global}$ into the multimodal token sequence to guide the frozen LLM in answer generation. Training updates only the soft prompts and linear projection parameters.}
\label{fig:main2}
\end{figure*}

GRASP partitions the visual token grid into spatial blocks and associates each block with a learnable soft prompt. It then computes question-conditioned sparse weights to aggregate block-wise prompts into a single global prompt, which is injected as an additional visual token to guide answer generation by a frozen language model. During training, supervision signals backpropagate only to the soft prompts and lightweight projection layers, while the vision and language backbones remain fixed, with the concept diagram comparing traditional SP and our approach shown in Fig.~\ref{fig:sp_concept}.
To clearly describe the workflow, we decompose GRASP into four components: \hyperref[sec:spatial_block_prompt]{(A)} Spatial Block-based Feature Encoding with Soft Prompts, \hyperref[sec:prompt_fusion]{(B)} Question-guided Sparse Soft Prompt Fusion, \hyperref[sec:text-generation]{(C)} Optimization Objective and Training Strategy, and \hyperref[sec:complexity]{(D)} Parameter and Computational Complexity. Fig.~\ref{fig:main2} provides an overview of the GRASP framework. 

\subsection{Spatial Block-based Feature Encoding with Soft Prompts}
\label{sec:spatial_block_prompt}
RS images usually exhibit high spatial resolution and complex structural layouts.
To capture region-specific local semantics while remaining compatible with large-scale vision--language models, we construct spatial block-level visual representations based on the visual token grid
produced by a frozen vision encoder, and introduce corresponding block-wise soft prompts
to bridge spatial visual information and language reasoning. Given an input image $x \in \mathbb{R}^{H \times W \times 3}$, where $H$ and $W$ denote the image height and width,
the frozen visual backbone encodes the image into $M = H_t \times W_t$ visual tokens
$\mathcal{T} = \{t_1, \dots, t_M\}$.
Each token $t_j \in \mathbb{R}^{d_v}$ represents a $d_v$-dimensional visual feature vector,
and all tokens are arranged in a two-dimensional grid of size $H_t \times W_t$,
preserving the spatial layout of the original image.
We uniformly partition the visual token grid into $N$ non-overlapping spatial blocks
$\mathcal{B} = \{B_1, \dots, B_N\}$, where each block $B_i \subset \mathcal{T}$
contains a set of spatially adjacent visual tokens.
Let $|B_i|$ denote the number of tokens in the $i$-th block.
We obtain the block-level visual representation by average pooling
the tokens within the block and adding a fixed positional encoding:
\begin{equation}
\label{eq:block_feature}
E_i = \frac{1}{|B_i|} \sum_{t_j \in B_i} t_j + \mathrm{PE}_i,
\end{equation}
where $E_i \in \mathbb{R}^{d_v}$ denotes the visual feature of the $i$-th spatial block,
and $\mathrm{PE}_i \in \mathbb{R}^{d_v}$ is a fixed positional encoding that represents the relative spatial location of the block.
Specifically, given $N$ spatial blocks, we introduce a set of block-wise soft prompts
$\{p_1, \dots, p_N\}$, where each $p_i \in \mathbb{R}^{d_p}$ is a $d_p$-dimensional trainable parameter
initialized from a zero-mean Gaussian distribution,
$p_i \sim \mathcal{N}(0, \sigma^2 I)$, with $\sigma^2$ denoting the initialization variance.
These soft prompts do not encode visual content directly.
Instead, they serve as task-level semantic adapters that model how information from different spatial regions
should influence the language model during multimodal reasoning.
By assigning distinct soft prompts to different spatial blocks, the model explicitly captures structural differences among regions within an image without modifying the frozen visual backbone. 

\subsection{Question-guided Sparse Soft Prompt Fusion}
\label{sec:prompt_fusion}

To selectively integrate block-level semantic cues, we propose a Question-guided Sparse Fusion mechanism.
This process operates at the input prompt construction stage and can be viewed as a decoupled query-guided selection mechanism. In this formulation, the visual spatial block features determine where to attend (region selection), while the learnable soft prompts specify what information to inject (content adaptation). Given an input question, we first extract a pooled textual representation $q \in \mathbb{R}^{d_t}$ by forwarding the question tokens through the frozen language model and mean-pooling the hidden states.
To compute relevance in a shared latent space, both the question representation and block-level visual features
are projected into a low-dimensional attention space of dimension $h$:
\begin{equation}
k_i = \operatorname{Proj}_K^{(v)}(E_i), \qquad
\tilde{q} = \operatorname{Proj}_Q^{(t)}(q),
\end{equation}
where $\operatorname{Proj}_K^{(v)} : \mathbb{R}^{d_v} \rightarrow \mathbb{R}^{h}$ and
$\operatorname{Proj}_Q^{(t)} : \mathbb{R}^{d_t} \rightarrow \mathbb{R}^{h}$
are learnable linear projections for visual and textual features, respectively.
We then measure the relevance of each spatial block via scaled dot-product similarity:
\begin{equation}
s_i = \frac{k_i^\top \tilde{q}}{\sqrt{h}}, \qquad i = 1, \dots, N.
\end{equation}
Here, $s_i$ measures the relevance between the question and the $i$-th spatial block.
The factor $\sqrt{h}$ scales the dot-product scores to stabilize their magnitude. To encourage sparsity and suppress irrelevant regions, the relevance scores are normalized using the Entmax~\cite{peters2019sparse} function:
\begin{equation}
\mathbf{w} = \operatorname{entmax}_{\alpha}(\mathbf{s}).
\end{equation}
Here, $\mathbf{w} = [w_1, \dots, w_N]^\top$ denotes the sparse importance weights. The hyperparameter $\alpha$ governs the sparsity level of Entmax, with $\alpha=1$ reducing to Softmax and larger values producing sparser weights. Finally, the block-wise soft prompts are aggregated into a single global prompt through a question-conditioned sparse weighting scheme:
\begin{equation}
\label{eq:prompt_fusion}
p_{\text{global}}
=
\sum_{i=1}^{N}
\left[
\operatorname{entmax}_{\alpha}
\left(
\frac{ \{ \operatorname{Proj}_K^{(v)}(E_j)
^\top \operatorname{Proj}_Q^{(t)}(q) \}_{j=1}^{N} }{\sqrt{h}}
\right)
\right]_i
\cdot p_i.
\end{equation}
In this unified formulation, $E_j$ refers to the spatial block representations defined in Eq.~\eqref{eq:block_feature}, while $p_{\text{global}} \in \mathbb{R}^{d_p}$ constitutes a task-specific parametric instruction summarizing question-relevant spatial regions (with $d_p = d_v$ to enable concatenation with visual tokens).
We inject the derived $p_{\text{global}}$ at the vision--language interface by constructing an extended visual input sequence as follows:
\begin{equation}
\label{eq:extended_visual_sequence}
\mathcal{T}' = [p_{\text{global}}, \mathcal{T}].
\end{equation}
The operator $[\cdot, \cdot]$ signifies concatenation along the sequence length dimension. The resulting spatially-enhanced sequence $\mathcal{T}'$ is then fed into the multimodal model, where it modulates the reasoning process.

\subsection{Optimization Objective and Training Strategy}
\label{sec:text-generation}

The language model performs autoregressive answer generation conditioned on the spatially-enhanced multimodal representation. Specifically, we concatenate the extended visual sequence $\mathcal{T}'$ (derived in Eq.~\eqref{eq:extended_visual_sequence}) with the tokenized question embeddings $X_Q$ to form the input context:
\begin{equation}
\mathcal{X}_{\text{in}} = [\mathcal{T}', X_Q].
\end{equation}
This combined sequence is fed into a frozen pretrained language decoder, which computes the conditional probability of the target answer $Y = \{y_1, \dots, y_L\}$. The optimization objective is to minimize the negative log-likelihood of the ground-truth tokens:
\begin{equation}
\mathcal{L}(\Theta) = - \sum_{t=1}^{L} \log P_{\text{LM}}(y_t \mid y_{<t}, \mathcal{X}_{\text{in}}; \Theta),
\end{equation}
where $y_{<t}$ denotes the preceding generated tokens.
Crucially, during backpropagation, gradients are computed solely with respect to the trainable parameter set $\Theta = \{ \{p_i\}_{i=1}^N, \operatorname{Proj}_K^{(v)}, \operatorname{Proj}_Q^{(t)} \}$. The massive parameters of the visual encoder and the language model remain frozen. This selective optimization ensures the model adapts to the RS domain through the lightweight soft prompts and fusion layers, maintaining parameter efficiency while preserving the generalization capability of the pretrained backbone.
\subsection{Parameter and Computational Complexity Analysis}
\label{sec:complexity}

Following the training setup described above, we analyze the parameter and computational complexity of GRASP.
Recall that the trainable parameter set is defined as $\Theta = \{ \{p_i\}_{i=1}^N, \operatorname{Proj}_K^{(v)}, \operatorname{Proj}_Q^{(t)} \}$. In the parameter analysis, we denote the learnable weight matrices corresponding to the projection layers as $\mathbf{W}_K \in \mathbb{R}^{h \times d_v}$ and $\mathbf{W}_Q^{(t)} \in \mathbb{R}^{h \times d_t}$, respectively.
To align multimodal semantics, GRASP projects the textual intent $q$ and block-level visual features $E_i$ into a shared low-rank latent space $\mathbb{R}^h$.
Consequently, the total learnable parameter count, $|\Theta_{\text{train}}|$, decomposes into:
\begin{equation}
\begin{aligned}
|\Theta_{\text{train}}| &= \text{Params}(\{p_i\}_{i=1}^N) + \text{Params}(\operatorname{Proj}_K^{(v)}) + \text{Params}(\operatorname{Proj}_Q^{(t)}) \\
&= \underbrace{N \times d_p}_{\text{Soft Prompts}} + \underbrace{(h \times d_v + h \times d_t)}_{\text{Projection Matrices}} \\
&= \mathcal{O}\left( N d_p + h(d_v + d_t) \right).
\end{aligned}
\end{equation}

Although the parameter count is primarily governed by the projection dimension $h$, the total size (typically $\approx$ 4M) remains negligible compared to the billions of parameters in the frozen backbone. This formulation reduces the number of trainable parameters and storage overhead. Regarding computational complexity, our method offers a dual advantage. First, the per-sample complexity of the fusion module $T_{\text{fusion}}$ is linear with respect to the number of blocks $N$:
\begin{equation}
\begin{aligned}
T_{\text{fusion}} &= T_{\text{proj}}(q) + \sum_{i=1}^{N} T_{\text{proj}}(E_i) + T_{\text{agg}}(p_{\text{global}}) \\
&= \mathcal{O}\big( N(d_v h + d_p) \big).
\end{aligned}
\end{equation}
Since $\mathcal{O}(T_{\text{fusion}})$ is several orders of magnitude smaller than the Transformer layers, the module introduces negligible latency. More importantly, GRASP introduces a highly efficient modulation mechanism for MLLM inference.
Instead of injecting a large number of task-specific tokens or modifying the original visual representations,
our fusion strategy produces a single global soft prompt token that summarizes question-relevant spatial regions, while retaining the original visual token sequence unchanged. This design introduces only a one-token overhead to the multimodal input, thereby preserving the inference efficiency of the pretrained language model while enabling spatially aware reasoning.
\begin{table*}[!t]
\scriptsize
\centering
\setlength{\tabcolsep}{3.5pt}
\renewcommand{\arraystretch}{1.3}
\caption{Unified results of LLaVA-Next on three benchmarks.
Params (M) denotes trainable parameters (millions).
Count, R/U, Pres., and Comp. correspond to different question categories, namely Counting, Rural/Urban, Presence, and Comparison. AA denotes Average Accuracy. \gtext{Green} values in parentheses denote the standard deviation over 3 independent runs. \textbf{Bold} and \underline{underline} mark the best and second-best results.}
\label{tab:unified_ablation_llava}
\begin{tabularx}{\textwidth}{
l |
c |
*{5}{>{\centering\arraybackslash}X} |
*{5}{>{\centering\arraybackslash}X} |
*{4}{>{\centering\arraybackslash}X}
}
\toprule
\multirow{2}{*}{\textbf{Method}} &
\multirow{2}{*}{\textbf{Params}} &
\multicolumn{5}{c|}{\textbf{RSVQA-LR}} &
\multicolumn{5}{c|}{\textbf{RSVQA-HR}} &
\multicolumn{4}{c}{\textbf{RSIVQA}} \\
\cmidrule(lr){3-7}
\cmidrule(lr){8-12}
\cmidrule(lr){13-16}
& & Count & R/U & Pres. & Comp. & AA
& Count & Area & Pres. & Comp. & AA
& Yes/No & Num. & Other & AA \\
\midrule
LLaVA-Next (Base)
& --
& 23.8 & 71.0 & 55.4 & 66.5 & 54.2
& 56.0 & 49.1 & 66.4 & 75.6 & 61.8
& 85.5 & 45.1 & 40.2 & 56.9 \\
\rowcolor{gray!20}
\multicolumn{16}{c}{\textbf{Additive parameter tuning methods}} \\
+ Adapter
& 16.7
& 28.0 \gtext{(±1.0)} & 81.9 \gtext{(±1.1)} & 74.6 \gtext{(±1.3)} & 77.1 \gtext{(±1.2)} & 65.4 \gtext{(±1.0)}
& 63.1 \gtext{(±1.2)} & \underline{68.8} \gtext{(±1.4)} & 78.4 \gtext{(±1.5)} & 80.9 \gtext{(±1.3)} & 72.8 \gtext{(±1.2)}
& 90.3 \gtext{(±1.1)} & 60.7 \gtext{(±1.6)} & 76.3 \gtext{(±1.8)} & 75.8 \gtext{(±1.4)} \\
+ LoRA
& 50.1
& 29.5 \gtext{(±0.9)} & 84.0 \gtext{(±1.0)} & 77.3 \gtext{(±1.1)} & \textbf{79.0} \gtext{(±1.0)} & 67.5 \gtext{(±0.9)}
& 64.2 \gtext{(±1.0)} & 68.6 \gtext{(±1.2)} & 78.6 \gtext{(±1.3)} & 81.8 \gtext{(±1.1)} & \underline{73.3} \gtext{(±1.0)}
& \underline{91.2} \gtext{(±0.9)} & 62.0 \gtext{(±1.4)} & 78.4 \gtext{(±1.6)} & 77.2 \gtext{(±1.2)} \\
+ DoRA
& 50.2
& 30.1 \gtext{(±0.8)} & \underline{84.3} \gtext{(±0.9)} & \underline{77.4} \gtext{(±1.0)} & 78.7 \gtext{(±0.9)} & \underline{67.6} \gtext{(±0.8)}
& \underline{64.8} \gtext{(±0.9)} & 68.5 \gtext{(±1.1)} & \underline{79.0} \gtext{(±1.2)} & \underline{82.1} \gtext{(±1.0)} & 73.6 \gtext{(±0.9)}
& \underline{91.2} \gtext{(±0.8)} & 62.7 \gtext{(±1.3)} & \underline{78.7} \gtext{(±1.5)} & \underline{77.5} \gtext{(±1.1)} \\
\rowcolor{gray!20}
\multicolumn{16}{c}{\textbf{Prompt-based tuning methods}} \\
+ Prompt Tuning
& 0.41
& 24.1 \gtext{(±1.1)} & 78.3 \gtext{(±1.3)} & 68.5 \gtext{(±1.5)} & 70.3 \gtext{(±1.4)} & 60.3 \gtext{(±1.2)}
& 57.4 \gtext{(±1.3)} & 53.9 \gtext{(±1.5)} & 70.2 \gtext{(±1.6)} & 76.7 \gtext{(±1.4)} & 64.6 \gtext{(±1.3)}
& 88.1 \gtext{(±1.2)} & 51.5 \gtext{(±1.7)} & 68.7 \gtext{(±1.9)} & 69.4 \gtext{(±1.5)} \\
+ APrompt
& 16.1
& 26.8 \gtext{(±1.0)} & 79.8 \gtext{(±1.2)} & 69.0 \gtext{(±1.4)} & 72.2 \gtext{(±1.3)} & 62.0 \gtext{(±1.1)}
& 58.6 \gtext{(±1.2)} & 57.4 \gtext{(±1.4)} & 71.9 \gtext{(±1.5)} & 77.2 \gtext{(±1.3)} & 66.3 \gtext{(±1.2)}
& 89.0 \gtext{(±1.1)} & 53.0 \gtext{(±1.6)} & 70.8 \gtext{(±1.8)} & 70.9 \gtext{(±1.4)} \\
+ VPT
& 6.70
& \underline{31.9} \gtext{(±0.9)} & 80.4 \gtext{(±1.0)} & 75.5 \gtext{(±1.1)} & 76.9 \gtext{(±1.0)} & 66.2 \gtext{(±0.9)}
& 61.5 \gtext{(±1.0)} & 67.3 \gtext{(±1.2)} & 78.5 \gtext{(±1.3)} & 80.1 \gtext{(±1.1)} & 71.9 \gtext{(±1.0)}
& 90.8 \gtext{(±0.9)} & \underline{62.9} \gtext{(±1.4)} & 75.4 \gtext{(±1.6)} & 76.4 \gtext{(±1.2)} \\
+ \textbf{GRASP (ours)}
& \textbf{4.25}
& \textbf{32.8} \gtext{(±0.8)} & \textbf{85.7} \gtext{(±0.9)} & \textbf{77.5} \gtext{(±1.0)} & \underline{78.8} \gtext{(±0.9)} & \textbf{68.7} \gtext{(±0.8)}
& \textbf{65.3} \gtext{(±0.9)} & \textbf{69.5} \gtext{(±1.1)} & \textbf{79.2} \gtext{(±1.2)} & \textbf{82.3} \gtext{(±1.0)} & \textbf{74.1} \gtext{(±0.9)}
& \textbf{92.4} \gtext{(±0.8)} & \textbf{63.6} \gtext{(±1.3)} & \textbf{78.9} \gtext{(±1.5)} & \textbf{78.3} \gtext{(±1.1)} \\
\bottomrule
\end{tabularx}
\end{table*}

\section{Experiments}
\subsection{Dataset Descriptions}

We evaluate the effectiveness of our proposed GRASP method using two widely adopted datasets: RSVQA~\cite{lobry2020rsvqa} and RSIVQA~\cite{zheng2021mutual}. The RSVQA dataset consists of two subsets: a low-resolution subset (Sentinel-2, 10m resolution, $256 \times 256$ pixels) and a high-resolution subset (USGS HRO, 15cm resolution, $512 \times 512$ pixels). The low-resolution subset contains 772 images with 77,232 IQA triplets, while the high-resolution subset includes 10,659 images and 1,066,316 IQA triplets. Questions in the dataset fall into four categories: Count, Presence, Comparison, and Rural/Urban, with the high-resolution subset also including the Area category.

The RSIVQA dataset contains 37,264 images and 111,693 IQA triplets, collected from various RS scene datasets such as UCM, Sydney, AID, HRRSD, and DOTA. The questions are categorized into three types: Yes/No, Number, and Others (e.g., spatial relation questions). Both datasets combine automated generation and manual annotation, providing a solid foundation for evaluating our method’s capabilities in global prompt structuring and local semantic modeling. 

In addition, we assess GRASP's generalizability by including the RSVLM-QA dataset~\cite{zi2025rsvlm}, which integrates multiple remote sensing datasets, including WHU Building, LoveDA, INRIA Aerial, and iSAID. RSVLM-QA contains 13,820 high-resolution images and 162,373 QA pairs, spanning five categories: Objects, Features, Spatial, Quantity, and Presence. Unlike RSVQA and RSIVQA, which rely on template-based question generation, RSVLM-QA employs a dual-track annotation pipeline combining GPT-4.1-driven descriptive QA and automated counting from segmentation labels, resulting in more open-ended and semantically rich questions.

For a fair comparison of model performance, we follow the 80\%/10\%/10\% split ratio used in~\cite{zheng2021mutual}, allocating 80\% of the images to the training set, 10\% to the validation set, and the remaining 10\% to the test set. The split is performed at the image level, and all QA pairs associated with the same image are assigned to the same subset, ensuring that no image appears across different subsets and avoiding data leakage.

\subsection{Experimental Details}
All experiments are conducted on a single NVIDIA RTX 3090 GPU. Unless otherwise specified, we use a shared optimization setting across the main RSVQA experiments and the additional analyses: the AdamW optimizer with an initial learning rate of $1 \times 10^{-4}$, weight decay of 0.01, linear warmup over the first 10\% of training steps followed by linear decay, batch size 6, training for up to 20 epochs, and early stopping with patience 5 based on validation performance. When a task requires a different evaluation protocol or output format, we keep the default training setup unchanged as much as possible and describe only the task-specific modifications in the corresponding subsections.

For the main controlled comparisons under the same backbone, all PEFT baselines, including LoRA, DoRA, VPT, APrompt, Prompt Tuning, and GRASP, are trained with the same data split, optimization schedule, model selection criterion, and backbone-specific preprocessing pipeline to ensure a fair comparison. For these main RSVQA experiments, each method is evaluated over three independent runs with different random seeds, and we report the mean and standard deviation across the three runs.

Image preprocessing follows the official recommendation of each backbone model. To maintain consistent spatial block partitioning in GRASP, images are resized with aspect ratio preserved and then center-padded to a fixed input size. For lower-resolution settings, we use $224 \times 224$, which helps control training and inference cost while maintaining a consistent block partition strategy across datasets with relatively low or varying native resolutions. For high-resolution settings, we use $512 \times 512$ to better preserve fine-grained spatial details, especially for small objects and local geometric structures. The specific choice for each dataset is described in the corresponding experimental subsection.

Our main experiments focus on two representative $7$B-scale vision-language models, LLaVA-Next-7B~\cite{liu2024llavanext} and Qwen2.5-VL-7B~\cite{bai2025qwen2}. In addition, we include Mini-InternVL-4B~\cite{gao2024mini} as a lightweight backbone in a supplementary scalability analysis. To ensure a fair comparison, we adhere strictly to the official preprocessing pipelines of each backbone and keep the vision encoders frozen during fine-tuning. The proposed GRASP uses a bottleneck dimension $h=512$, with the soft prompt dimension aligned to the language model hidden size and an Entmax sparsity parameter $\alpha=1.5$. The spatial block count $N$ is treated as a key hyperparameter and selected through ablation studies (see Section~\ref{sec:ablation}). In the main RSVQA experiments, we set $N=4$ for RSVQA-LR and RSIVQA, and $N=16$ for RSVQA-HR.
\begin{table*}[!t]
\scriptsize
\centering
\setlength{\tabcolsep}{3.5pt}
\renewcommand{\arraystretch}{1.3}
\caption{Unified results of Qwen2.5-VL-7B on three benchmarks.
Params (M) denotes trainable parameters (millions).
The abbreviations follow the same definitions as in Table~\ref{tab:unified_ablation_llava}.
\gtext{Green} values in parentheses denote the standard deviation over 3 independent runs.
\textbf{Bold} and \underline{underline} mark the best and second-best results.}
\label{tab:unified_ablation_qwen}
\begin{tabularx}{\textwidth}{
l |
c |
*{5}{>{\centering\arraybackslash}X} |
*{5}{>{\centering\arraybackslash}X} |
*{4}{>{\centering\arraybackslash}X}
}
\toprule

\multirow{2}{*}{\textbf{Method}} &
\multirow{2}{*}{\textbf{Params}} &
\multicolumn{5}{c|}{\textbf{RSVQA-LR}} &
\multicolumn{5}{c|}{\textbf{RSVQA-HR}} &
\multicolumn{4}{c}{\textbf{RSIVQA}} \\
\cmidrule(lr){3-7}
\cmidrule(lr){8-12}
\cmidrule(lr){13-16}

& & Count & R/U & Pres. & Comp. & AA
& Count & Area & Pres. & Comp. & AA
& Yes/No & Num. & Other & AA \\
\midrule

Qwen2.5-VL (Base)
& --
& 20.9 & 66.0 & 62.3 & 71.4 & 55.1 
& 52.6 & 61.4 & 65.6 & 76.4 & 64.0 
& 87.3 & 43.8 & 38.7 & 56.6 \\

\rowcolor{gray!20}
\multicolumn{16}{c}{\textbf{Additive parameter tuning methods}} \\

+ Adapter
& 16.7
& 27.6 \gtext{(±1.1)} & 81.4 \gtext{(±1.1)} & 79.0 \gtext{(±1.2)} & 80.5 \gtext{(±1.2)} & 67.1 \gtext{(±1.0)}
& 60.3 \gtext{(±1.3)} & 69.7 \gtext{(±1.4)} & 66.4 \gtext{(±1.5)} & 80.3 \gtext{(±1.2)} & 69.2 \gtext{(±1.3)}
& 90.5 \gtext{(±1.1)} & 58.0 \gtext{(±1.5)} & 82.8 \gtext{(±1.7)} & 77.1 \gtext{(±1.4)} \\

+ LoRA
& 50.1
& 29.5 \gtext{(±0.9)} & \textbf{83.0} \gtext{(±1.0)} & 80.8 \gtext{(±1.1)} & \underline{81.3} \gtext{(±1.0)} & 68.7 \gtext{(±0.9)}
& 62.1 \gtext{(±1.1)} & 70.6 \gtext{(±1.2)} & 70.3 \gtext{(±1.3)} & 81.9 \gtext{(±1.1)} & 71.2 \gtext{(±1.0)}
& \underline{92.0} \gtext{(±0.9)} & 60.9 \gtext{(±1.4)} & \textbf{89.2} \gtext{(±1.6)} & \underline{80.7} \gtext{(±1.2)} \\

+ DoRA 
& 50.2
& 30.0 \gtext{(±0.8)} & \underline{82.9} \gtext{(±0.9)} & \underline{81.2} \gtext{(±1.0)} & 81.1 \gtext{(±0.9)} & \underline{68.8} \gtext{(±0.8)}
& \underline{62.4} \gtext{(±1.0)} & \underline{70.8} \gtext{(±1.1)} & \underline{71.0} \gtext{(±1.2)} & \underline{82.1} \gtext{(±1.0)} & \underline{71.6} \gtext{(±0.9)}
& 91.7 \gtext{(±0.8)} & 61.1 \gtext{(±1.3)} & \underline{89.0} \gtext{(±1.5)} & 80.6 \gtext{(±1.1)} \\

\rowcolor{gray!20}
\multicolumn{16}{c}{\textbf{Prompt-based tuning methods}} \\

+ Prompt Tuning
& 0.41
& 22.3 \gtext{(±1.2)} & 77.8 \gtext{(±1.4)} & 71.4 \gtext{(±1.5)} & 72.9 \gtext{(±1.4)} & 61.1 \gtext{(±1.2)}
& 54.1 \gtext{(±1.4)} & 63.0 \gtext{(±1.6)} & 67.1 \gtext{(±1.5)} & 78.5 \gtext{(±1.4)} & 65.7 \gtext{(±1.3)}
& 89.3 \gtext{(±1.2)} & 50.2 \gtext{(±1.8)} & 75.1 \gtext{(±2.0)} & 71.5 \gtext{(±1.5)} \\

+ APrompt
& 16.1
& 23.4 \gtext{(±1.1)} & 78.1 \gtext{(±1.3)} & 72.5 \gtext{(±1.4)} & 73.0 \gtext{(±1.3)} & 61.8 \gtext{(±1.1)}
& 56.3 \gtext{(±1.3)} & 64.2 \gtext{(±1.5)} & 67.4 \gtext{(±1.4)} & 79.4 \gtext{(±1.3)} & 66.8 \gtext{(±1.2)}
& 90.4 \gtext{(±1.1)} & 51.6 \gtext{(±1.7)} & 77.4 \gtext{(±1.9)} & 73.1 \gtext{(±1.4)} \\

+ VPT
& 6.70
& \underline{30.3} \gtext{(±1.0)} & 80.9 \gtext{(±1.1)} & 79.3 \gtext{(±1.1)} & 79.3 \gtext{(±1.0)} & 67.5 \gtext{(±0.9)}
& 61.2 \gtext{(±1.1)} & 68.4 \gtext{(±1.2)} & 68.2 \gtext{(±1.3)} & 80.5 \gtext{(±1.1)} & 69.6 \gtext{(±1.1)}
& 91.2 \gtext{(±1.0)} & \underline{61.2} \gtext{(±1.5)} & 81.4 \gtext{(±1.7)} & 77.9 \gtext{(±1.3)} \\

+ \textbf{GRASP (ours)}
& \textbf{4.25}
& \textbf{31.1} \gtext{(±0.8)} & 82.6 \gtext{(±1.0)} & \textbf{81.6} \gtext{(±1.0)} & \textbf{81.5} \gtext{(±0.9)} & \textbf{69.2} \gtext{(±0.8)}
& \textbf{62.7} \gtext{(±0.9)} & \textbf{71.0} \gtext{(±1.0)} & \textbf{73.9} \gtext{(±1.1)} & \textbf{82.5} \gtext{(±0.9)} & \textbf{72.5} \gtext{(±0.8)}
& \textbf{93.1} \gtext{(±0.8)} & \textbf{61.5} \gtext{(±1.3)} & \underline{89.0} \gtext{(±1.4)} & \textbf{81.2} \gtext{(±1.1)} \\

\bottomrule
\end{tabularx}
\end{table*}

For the main RSVQA benchmarks, following the standard evaluation protocol of RSVQA~\cite{lobry2020rsvqa}, we report Accuracy based on exact matching between the predicted answer and the ground truth. To bridge the gap between free-form text generation in MLLMs and the fixed-vocabulary answer space of the datasets, the generated text is normalized by lowercasing, removing punctuation and extra whitespace, and converting number words to their numeric forms (e.g., ``four'' to ``4''). The final answer is extracted from the normalized output according to the corresponding answer type, such as a number for numerical questions or a keyword for categorical questions, and compared with the ground truth using exact matching. All compared methods follow the same post-processing procedure. Task-specific evaluation metrics for the supplementary experiments are described in their respective subsections.

For a fair comparison, all baseline methods keep the vision encoder frozen. VPT~\cite{jia2022visual} introduces learnable visual prompt tokens into the visual Transformer while maintaining fixed backbone weights. Adapter~\cite{houlsby2019parameter} (hidden dimension = 64) is applied to the Transformer layers of the language model. LoRA~\cite{hu2022lora} (rank = 64) and DoRA~\cite{liu2024dora} (rank = 64) are both applied to the same set of linear projections in the language model Transformer for a controlled comparison. Prompt Tuning~\cite{lester2021power} and APrompt~\cite{wang2023aprompt} inject prompt tokens at the input or intermediate layers. This setting allows performance differences to primarily reflect the effectiveness of different adaptation strategies.

\begin{table}[t]
\centering
\caption{VQA performance on RSVLM-QA. All answer correctness is judged automatically by GPT-4.1. Avg. denotes the average accuracy over the five VQA categories. \gtext{Green} values in parentheses denote the standard deviation over 3 independent runs. \textbf{Bold} and \underline{underline} mark the best and second-best results.}
\label{tab:rsvlmqa_vqa01}
\scriptsize
\setlength{\tabcolsep}{2.6pt} 
\renewcommand{\arraystretch}{2} 
\begin{tabular}{lcccccc}
\toprule
\textbf{Method} & \textbf{Objects} & \textbf{Features} & \textbf{Spatial} & \textbf{Quantity} & \textbf{Presence} & \textbf{Avg.} \\
\midrule
Qwen2.5-VL (Base) & 58.6 & 74.2 & 57.9 & 28.3 & 46.3 & 53.1 \\

\rowcolor{gray!20}
\multicolumn{7}{c}{\textbf{Additive parameter tuning methods}} \\
+ Adapter & \makecell{63.2 \\ \gtext{(±1.1)}} & \makecell{77.3 \\ \gtext{(±1.2)}} & \makecell{62.4 \\ \gtext{(±1.4)}} & \makecell{34.2 \\ \gtext{(±1.5)}} & \makecell{54.1 \\ \gtext{(±1.2)}} & \makecell{58.2 \\ \gtext{(±0.9)}} \\
+ LoRA    & \makecell{71.4 \\ \gtext{(±0.8)}} & \makecell{\underline{80.1} \\ \gtext{(±1.0)}} & \makecell{65.8 \\ \gtext{(±1.1)}} & \makecell{36.1 \\ \gtext{(±1.3)}} & \makecell{\underline{68.4} \\ \gtext{(±0.9)}} & \makecell{64.4 \\ \gtext{(±0.7)}} \\
+ DoRA    & \makecell{\textbf{72.2} \\ \gtext{(±0.7)}} & \makecell{79.3 \\ \gtext{(±0.9)}} & \makecell{\underline{67.9} \\ \gtext{(±1.0)}} & \makecell{\underline{37.2} \\ \gtext{(±1.2)}} & \makecell{67.8 \\ \gtext{(±1.0)}} & \makecell{\underline{64.9} \\ \gtext{(±0.6)}} \\

\rowcolor{gray!20}
\multicolumn{7}{c}{\textbf{Prompt-based tuning methods}} \\
+ Prompt Tuning & \makecell{62.3 \\ \gtext{(±1.3)}} & \makecell{76.1 \\ \gtext{(±1.4)}} & \makecell{61.4 \\ \gtext{(±1.6)}} & \makecell{30.8 \\ \gtext{(±1.8)}} & \makecell{52.7 \\ \gtext{(±1.5)}} & \makecell{56.7 \\ \gtext{(±1.2)}} \\
+ APrompt       & \makecell{64.8 \\ \gtext{(±1.2)}} & \makecell{78.5 \\ \gtext{(±1.1)}} & \makecell{63.2 \\ \gtext{(±1.3)}} & \makecell{32.5 \\ \gtext{(±1.4)}} & \makecell{55.9 \\ \gtext{(±1.2)}} & \makecell{59.0 \\ \gtext{(±0.9)}} \\
+ VPT           & \makecell{67.1 \\ \gtext{(±1.0)}} & \makecell{79.4 \\ \gtext{(±1.0)}} & \makecell{66.0 \\ \gtext{(±1.2)}} & \makecell{34.7 \\ \gtext{(±1.3)}} & \makecell{61.8 \\ \gtext{(±1.1)}} & \makecell{61.8 \\ \gtext{(±0.8)}} \\

\textbf{+ GRASP (ours)} & \makecell{\underline{72.0} \\ \gtext{(±0.8)}} & \makecell{\textbf{80.9} \\ \gtext{(±0.9)}} & \makecell{\textbf{70.7} \\ \gtext{(±1.1)}} & \makecell{\textbf{38.3} \\ \gtext{(±1.3)}} & \makecell{\textbf{72.1} \\ \gtext{(±0.9)}} & \makecell{\textbf{66.8} \\ \gtext{(±0.7)}} \\
\bottomrule
\end{tabular}
\end{table}

\begin{figure}[!t] 
  \centering
  \includegraphics[width=0.95\linewidth]{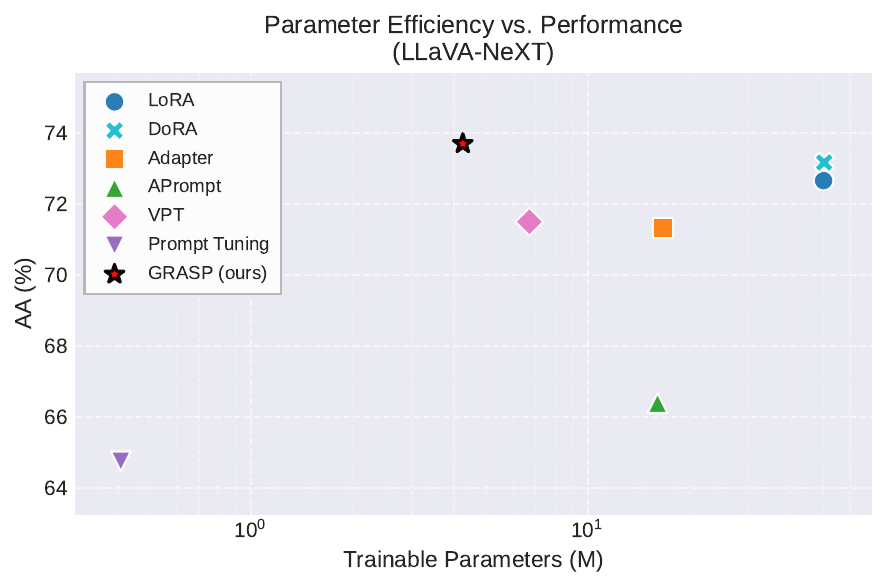}
  \caption{Parameter efficiency vs. performance on LLaVA-Next. The y-axis denotes the overall accuracy (mean of the AA scores on RSVQA-LR, RSVQA-HR, and RSIVQA), plotted against trainable parameters (log scale). Our method, GRASP, is represented by a red star. Note that proximity to the top-left corner corresponds to higher accuracy with lower parameter costs.}
  \label{fig:llava_performance}
\end{figure}

\subsection{Performance Evaluation}
\begin{table}[h]
\centering
\caption{Reference comparison with representative RS-MLLMs using publicly reported RSVQA results. The AA values in this table are calculated only over the question categories shared by the compared models and are therefore not directly equivalent to the AA values reported in the main experimental tables. GRASP results are obtained with the LLaVA-NeXT-7B backbone. \textcolor{gray}{Gray-colored} numbers denote zero-shot results; ``---'' indicates that the metric is not provided. This table is presented as a reference comparison rather than a strictly controlled head-to-head evaluation.}
\label{tab:rsvqa_metric_ablation02}
\scriptsize
\setlength{\tabcolsep}{3.5pt}
\renewcommand{\arraystretch}{1.1}
\begin{tabular}{l | c c c c | c c c}
\toprule
\multirow{2}{*}{\textbf{Model}} 
& \multicolumn{4}{c|}{\textbf{RSVQA-LR}} 
& \multicolumn{3}{c}{\textbf{RSVQA-HR}} \\
& \textbf{R/U} & \textbf{Pres.} & \textbf{Comp.} & \textbf{AA (\%)} 
& \textbf{Pres.} & \textbf{Comp.} & \textbf{AA (\%)} \\
\midrule
GeoChat      
& 91.0 & 90.3 & 94.0 & 91.7 
& \textcolor{gray}{58.4} 
& \textcolor{gray}{83.19} 
& \textcolor{gray}{72.3} \\
EarthGPT     
& --- & --- & --- & ---  
& \textcolor{gray}{62.7} 
& \textcolor{gray}{79.53} 
& \textcolor{gray}{72.0} \\
SkyEyeGPT & 88.9 & 88.6 & 75.0 & 84.1 & 80.0 & 80.1 &80.0\\
GRASP (Ours) 
& 85.7 & 77.5 & 78.8 & 80.6  
& 79.2 & 82.3 & 80.8 \\
\bottomrule
\end{tabular}
\end{table}
\textbf{Comparison with General PEFT Methods.} In this section, we evaluate GRASP against existing fine-tuning paradigms and examine the trade-off between reasoning accuracy and parameter efficiency across two representative MLLM backbones, namely LLaVA-Next and Qwen2.5-VL. As shown in Tables~\ref{tab:unified_ablation_llava} and~\ref{tab:unified_ablation_qwen}, GRASP demonstrates competitive performance on all benchmarks while maintaining a relatively small parameter footprint. Taking Qwen2.5-VL as a representative example, GRASP achieves 72.5\% average accuracy (AA) on RSVQA-HR, yielding a modest gain over LoRA (71.2\%) and slightly higher AA than its enhanced variant DoRA (71.6\%). Importantly, GRASP attains this improvement with lower computational overhead: while LoRA updates 50.1M parameters (0.72\% of the backbone), GRASP requires only 4.25M trainable parameters (0.06\%), corresponding to an $11.8\times$ reduction in trainable weights. This consistent performance gap across both backbones suggests a key difference in adaptation mechanisms. While LoRA applies low-rank updates globally to model weights, GRASP employs a spatially explicit prompting strategy that associates learnable soft prompts with specific image regions. This region-aware design better preserves localized semantic information, enabling more effective capture of spatial nuances in RS imagery. 
\begin{figure*}[!t]
    \centering
    \includegraphics[width=1\linewidth,trim=0 15mm 0 15mm,clip]{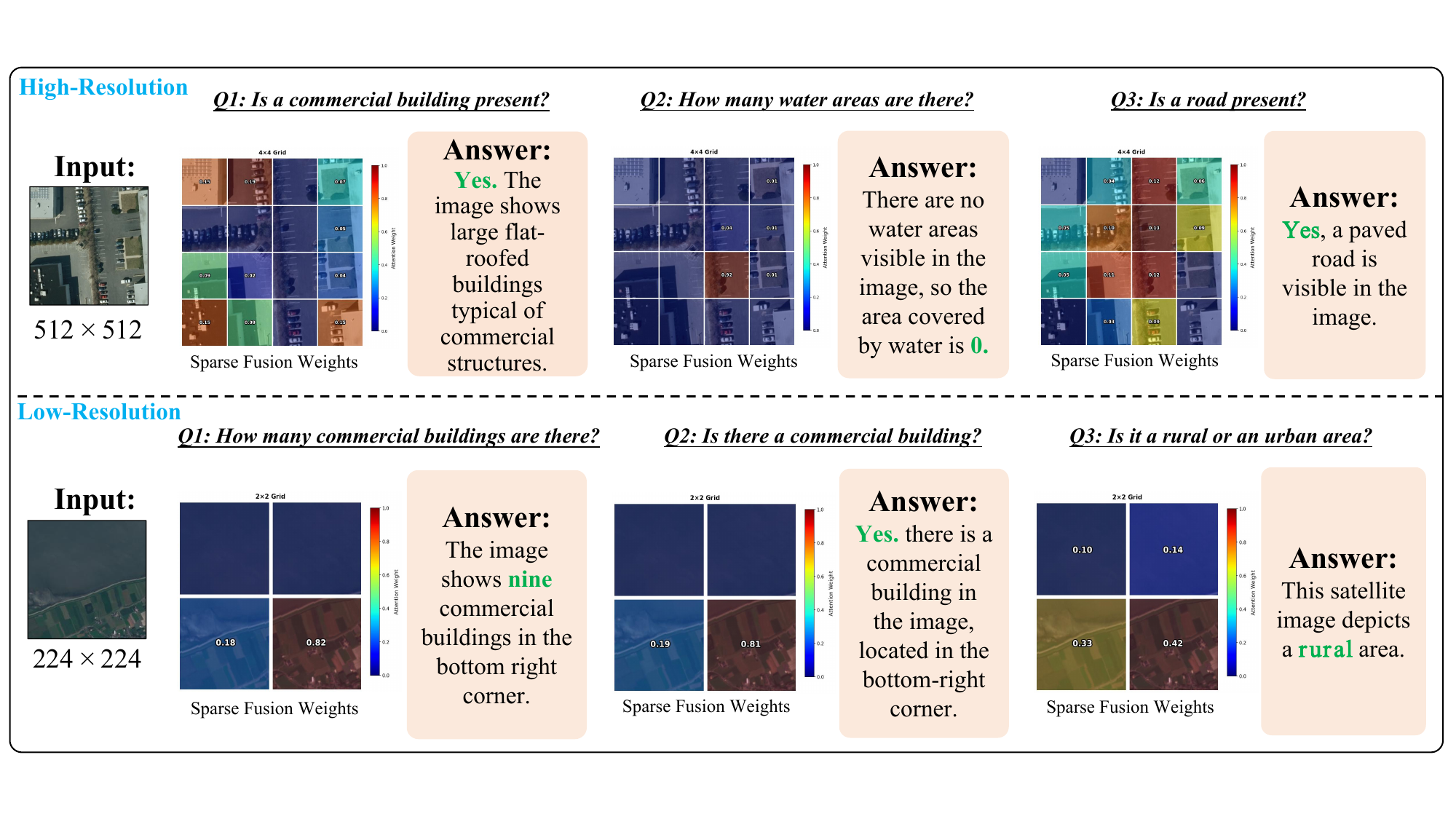}
    \caption{\textbf{Visualization of Question-guided Sparse Fusion Weights across different input resolutions.} The top row shows the $4 \times 4$ grid weights for the $512 \times 512$ input, while the bottom row displays the $2 \times 2$ grid for the $224 \times 224$ input. These weights represent the contribution of each spatial block to the final representation, with answers highlighted in green denoting correct predictions. The displayed answers are raw free-form model outputs.}
    \label{fig:exp_main}
\end{figure*}

Furthermore, compared with VPT, a representative prompt-based method, GRASP improves accuracy by 2.9\% on RSVQA-HR and 1.7\% on RSVQA-LR when using the Qwen2.5-VL backbone. These gains support the hypothesis that the global prompts used in standard prompt tuning are insufficient for spatially dispersed targets in RS imagery. By decomposing visual cues into region-aware sparse prompts and aggregating them through a question-guided fusion mechanism, GRASP effectively filters irrelevant background information and enhances region-specific spatial semantic modeling. This advantage is further illustrated in Fig.~\ref{fig:qwen_performance} and~\ref{fig:llava_performance}, which visualizes the trade-off between parameter efficiency and performance. In the scatter plot, GRASP (marked with a red star) occupies the upper-left region, indicating a favorable balance between accuracy and parameter cost. 

Beyond the overall improvements, the gains brought by GRASP are not uniformly distributed across all question categories. We observe that the improvement is generally more evident on Presence and Comparison questions, where correct answering depends strongly on identifying question-relevant local regions while suppressing distracting background content. This is consistent with the design of GRASP, which explicitly associates learnable prompts with spatial blocks and selectively aggregates them through question-guided sparse fusion. By contrast, the gains on Count-related questions are relatively smaller. We conjecture that counting requires stronger global spatial consistency and more complete aggregation across multiple instances, whereas fixed spatial partitioning may fragment relevant evidence when objects are highly sparse or distributed across block boundaries. As a result, while GRASP is particularly effective for question types that rely on precise region selection and background suppression, its benefit is comparatively more limited for tasks requiring more global quantity estimation. These observations further suggest that the gains of GRASP are closely related to its ability to handle sparse targets and suppress dominant background regions in remote sensing imagery.

We extend our evaluation to the RSVLM-QA dataset to assess GRASP's generalization across more diverse and challenging scenarios. As shown in Table~\ref{tab:rsvlmqa_vqa01}, GRASP performs well on the RSVLM-QA dataset, with consistent improvements in both the Presence and Spatial categories. These gains stem from GRASP's region-aware sparse prompting and question-guided fusion mechanisms, which effectively suppress background noise and focus on task-relevant regions. This approach offers a clear advantage in tasks where spatial attention plays a critical role.

However, in the Objects category, GRASP slightly underperforms DoRA, the strongest method. This may arise because the Objects category typically requires the model to recognize and distinguish a wide variety of targets, whereas GRASP's sparsity can lead to insufficient representation of low-salience or secondary targets. For the Quantity category, we consider it the most challenging aspect of remote sensing VQA tasks. Although GRASP's question-guided sparse fusion mechanism effectively identifies discriminative regions, when targets are numerous or unevenly distributed, fixed grid partitioning can fragment objects of the same class across multiple blocks. This fragmentation affects the accuracy and completeness of quantity estimation.

\begin{figure*}[t] 
  \centering
  \includegraphics[width=1\linewidth]{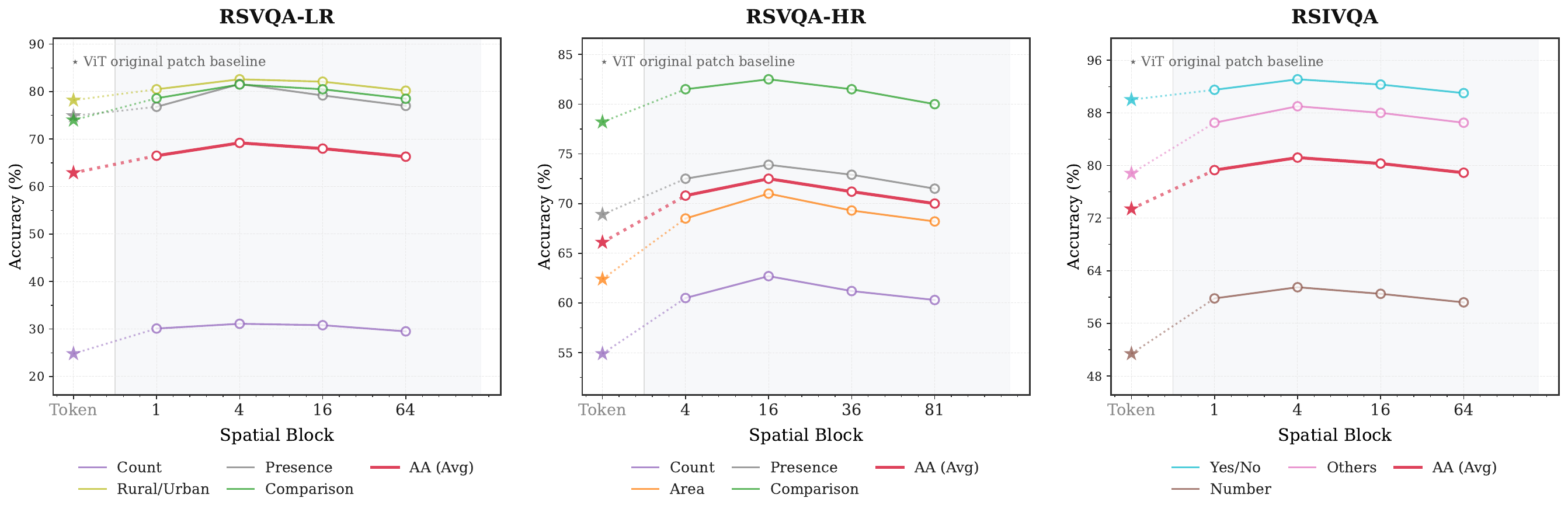}
  \caption{Impact of spatial block count ($N$) on performance with the Qwen2.5-VL-7B backbone across RSVQA-LR, RSVQA-HR, and RSIVQA datasets. The 'Token' marker ($\star$) denotes the original patch-level baseline, while the solid lines track the performance trend under our proposed spatial blocking strategy.}
  \label{fig:ablation_blocks}
\end{figure*}
\textbf{Comparison with Remote Sensing-Specific MLLMs.} To provide a broader perspective on GRASP's positioning within the remote sensing domain, we present a reference comparison with several representative remote sensing-adapted MLLMs (GeoChat, EarthGPT, SkyEyeGPT) using publicly reported results on the RSVQA benchmarks, as shown in Table~\ref{tab:rsvqa_metric_ablation02}.

Note that these models are typically trained via large-scale domain-specific instruction tuning on extensive remote sensing datasets, whereas GRASP focuses on lightweight, parameter-efficient adaptation of general-purpose backbones with frozen weights and limited supervision. Direct head-to-head comparison is therefore not strictly controlled due to differences in backbone architectures, training protocols, and evaluation settings (zero-shot and fine-tuned settings). Nevertheless, from this reference perspective, GRASP achieves an AA of 80.8\% on RSVQA-HR while using a parameter-efficient adaptation strategy, highlighting its potential as a resource-friendly alternative for remote sensing visual question answering.

\subsection{Qualitative Results}
To provide intuitive insights into how GRASP performs cross-modal reasoning, we visualize representative success cases and their corresponding attention heatmaps in Fig.~\ref{fig:exp_main}. These heatmaps are obtained by projecting the learned question-guided sparse fusion weights onto the original image spatial blocks, illustrating how the model allocates spatial attention under different queries.

As shown in the high-resolution examples, GRASP exhibits clear question-dependent spatial activation patterns. For instance, when answering whether a commercial building is present (Q1), the model assigns higher weights to spatial blocks covering large flat-roofed structures, while assigning lower weights to surrounding regions. In object counting queries (Q2), the activated blocks correspond to multiple distinct building instances, indicating that the model attends to individual objects relevant to the counting task. Similarly, for road-related questions (Q3), GRASP emphasizes elongated linear regions that align with paved road structures in the image. A similar behavior can be observed under low-resolution inputs, where GRASP selectively focuses on a small subset of spatial blocks associated with the queried semantic concepts. Despite reduced visual detail, the sparse fusion weights remain concentrated on regions that are most informative for the given question, suggesting that the spatially explicit prompting mechanism facilitates localized visual reasoning across different input resolutions.

\begin{figure}[t] 
  \centering
  \includegraphics[width=1\linewidth]{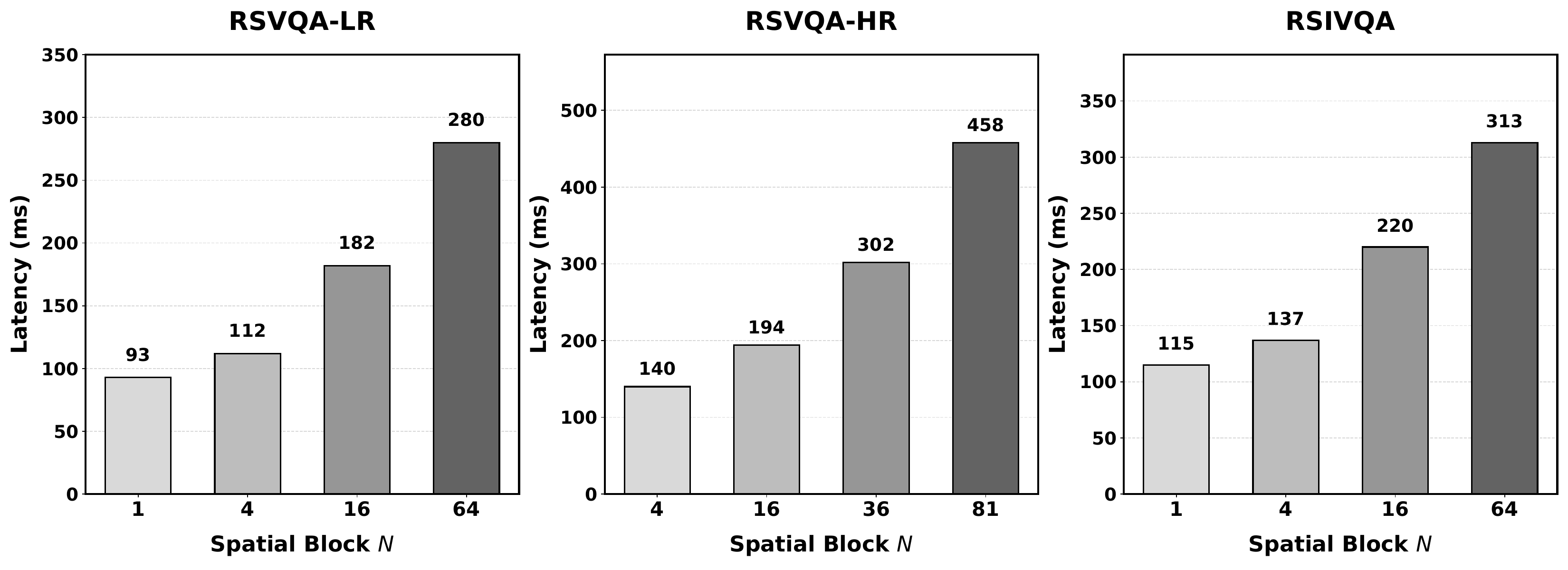}
  \caption{Inference latency versus the number of spatial blocks $N$ on RSVQA-LR, RSVQA-HR, and RSIVQA. Latency is reported as the average inference time per sample.}
  \label{fig:ablation_blocks01}
\end{figure}
\subsection{Ablation Study}
\label{sec:ablation}
In this section, we conduct a comprehensive ablation study to evaluate the impact of key architectural hyperparameters on the performance of GRASP. Specifically, we focus on two critical factors governing the model's spatial reasoning capability: the granularity of spatial block partitioning ($N$) and the sparsity level of the Entmax-based prompt fusion mechanism ($\alpha$). To maintain conciseness while ensuring representative analysis, we report all ablation results using the Qwen2.5-VL-7B backbone.

\begin{figure}[h] 
  \centering
  \includegraphics[width=1\linewidth]{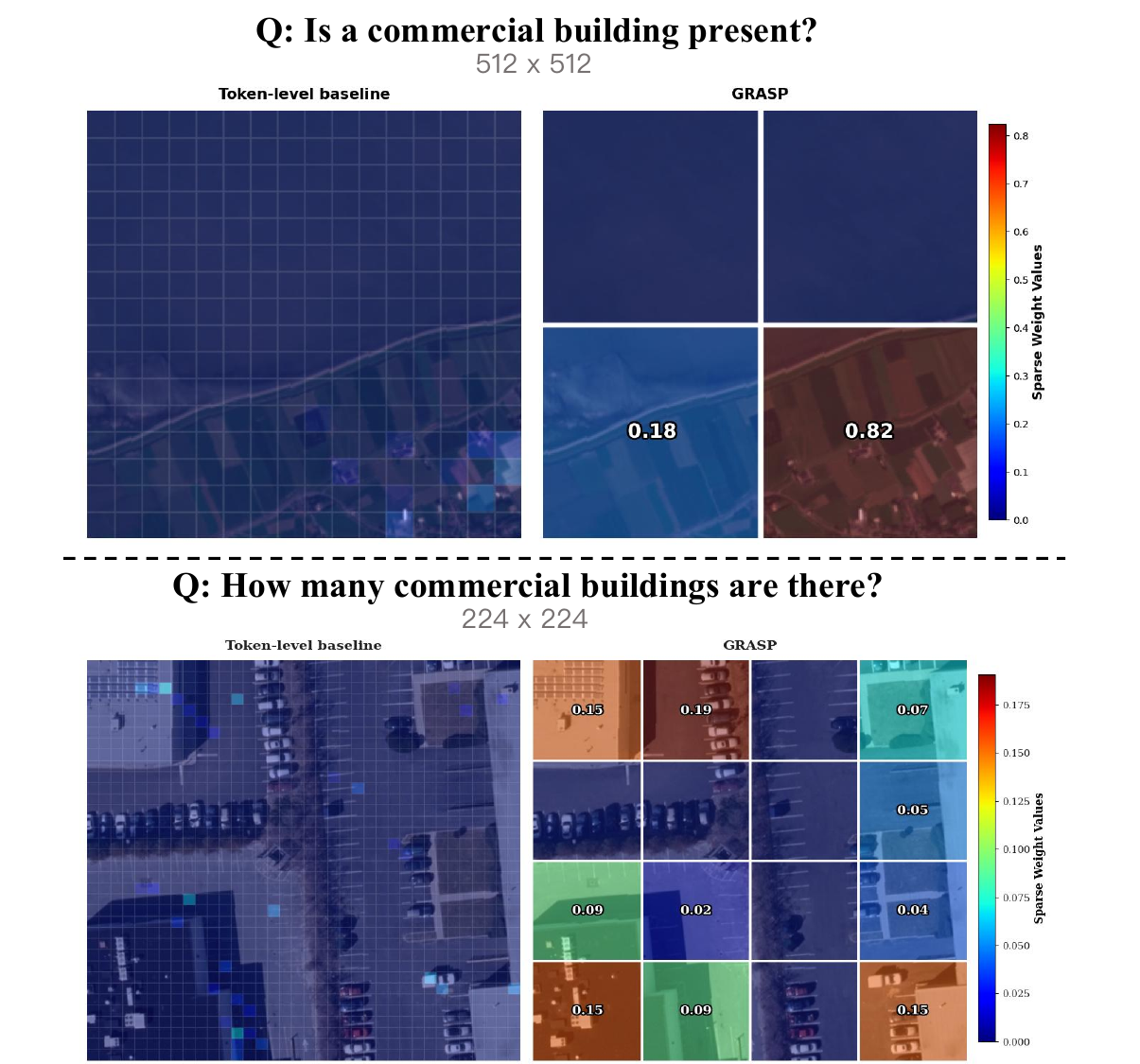}
  \caption{Visualization of sparse weight distributions under different input resolutions. The two rows present the token-level baseline and our GRASP across $512\times512$ and $224\times224$ scales, respectively. For visual clarity, weight magnitudes are independently normalized to their respective local maximums.}
  \label{fig:weight}
\end{figure}

\textbf{Number of Spatial Blocks.}
To evaluate the role of block-level spatial grouping in GRASP, we construct a Token-level Question-Guided Weighting baseline, which directly performs question-guided similarity computation and sparse weighting on the original ViT patch tokens without introducing any spatial region aggregation. As shown in Fig.~\ref{fig:ablation_blocks}, the token-level baseline yields moderate improvements over conventional soft prompting on several metrics, indicating that question-guided token-level interactions are themselves beneficial to some extent. However, its weight responses remain relatively scattered (see Fig.~\ref{fig:weight}) and are more sensitive to local texture variations, making it difficult to form stable and coherent region-level response patterns.

Further analysis of the performance variations under different spatial granularities $N$ reveals that the optimal spatial granularity is closely related to the spatial resolution of the input image and the density of target distributions. For scenarios with lower image resolutions or relatively larger target scales, a coarser partition ($N=4$, i.e., a $2\times2$ grid) generally achieves better performance. When the partition is further refined to $N=16$ or $64$, performance begins to degrade, which may be attributed to excessive fragmentation of regional structures and weakened local semantic coherence. In contrast, for high-resolution scenarios with denser target distributions, a finer-grained partition ($N=16$, i.e., a $4\times4$ grid) performs better, while excessively coarse or overly fine partitions both lead to performance degradation.

By jointly considering the token-level baseline, the performance variations under different $N$, and the weight visualization results, a consistent observation can be made. Token-level question-guided weighting struggles to stably model region-level semantics without explicit spatial organization, while introducing appropriate block-level grouping enables more stable region-level modeling across multiple datasets. These findings suggest that block-level spatial grouping introduces an effective spatial organizational constraint over patch tokens, thereby helping improve the stability and consistency of region-aware prompt adaptation. Such structured modeling is particularly well suited to remote sensing scenarios characterized by sparse targets, background-dominated scenes, and task reasoning that relies on localized regions.

\begin{table*}[t]
\centering
\scriptsize
\setlength{\tabcolsep}{4.5pt}
\renewcommand{\arraystretch}{0.8}
\caption{
Unified ablation on Entmax sparsity parameter $\alpha$ for GRASP on Qwen2.5-VL-7B across RSVQA-LR, RSVQA-HR, and RSIVQA datasets.
$\alpha=1.0$ and $\alpha=2.0$ correspond to Softmax and Sparsemax, respectively.
Results are reported as accuracy (\%). The block number $N$ is fixed to the best setting from the block-number ablation (RSVQA-LR/RSIVQA: $N=4$; RSVQA-HR: $N=16$).
}
\label{tab:ablation_alpha_unified}
\begin{tabularx}{\textwidth}{
c |
*{5}{>{\centering\arraybackslash}X} |
*{5}{>{\centering\arraybackslash}X} |
*{4}{>{\centering\arraybackslash}X}
}
\toprule

\multirow{2}{*}{$\alpha$} &
\multicolumn{5}{c|}{\textbf{RSVQA-LR}} &
\multicolumn{5}{c|}{\textbf{RSVQA-HR}} &
\multicolumn{4}{c}{\textbf{RSIVQA}} \\
\cmidrule(lr){2-6}
\cmidrule(lr){7-11}
\cmidrule(lr){12-15}

& Count & R/U & Pres. & Comp. & AA
& Count & Area & Pres. & Comp. & AA
& Yes/No & Num. & Other & AA \\
\midrule

1.0
& 28.4 & 79.2 & 78.0 & 78.1 & 65.9
& 59.5 & 68.0 & 70.5 & 78.8 & 69.2
& 91.0 & 58.7 & 82.6 & 77.4 \\

1.2
& 29.6 & 81.0 & 79.7 & 79.9 & 67.6
& 61.1 & 69.4 & 72.0 & 80.5 & 70.8
& 91.8 & 60.0 & 86.2 & 79.3 \\

\textbf{1.5}
& \textbf{31.1} & \textbf{82.6} & \textbf{81.6} & \textbf{81.5} & \textbf{69.2}
& \textbf{62.7} & \textbf{71.0} & \textbf{73.9} & \textbf{82.5} & \textbf{72.5}
& \textbf{93.1} & \textbf{61.5} & \textbf{89.0} & \textbf{81.2} \\

1.8
& 30.5 & 81.5 & 80.5 & 80.6 & 68.3
& 61.8 & 70.1 & 72.5 & 81.2 & 71.4
& 92.6 & 60.7 & 88.1 & 80.5 \\

2.0
& 29.3 & 80.1 & 78.9 & 79.3 & 66.9
& 60.5 & 68.7 & 71.4 & 79.9 & 70.1
& 92.2 & 60.3 & 86.8 & 79.8 \\

\bottomrule
\end{tabularx}
\end{table*}

\textbf{Impact of Spatial Block Count on Inference Latency.}
In addition to its effect on accuracy, we further examine the relationship between the spatial block count $N$ and inference latency. As illustrated in Fig.~\ref{fig:ablation_blocks01}, the inference latency increases monotonically with $N$ across all three datasets. This trend reflects the escalating computational overhead associated with finer-grained grid partitioning, as a larger number of blocks necessitates more extensive feature aggregation and question-guided fusion operations.

The increase is particularly noticeable on RSVQA-HR, where the input resolution is higher and the corresponding visual representation is more detailed. For example, the latency rises from 140 ms at $N=4$ to 458 ms at $N=81$, indicating that overly fine partitioning can moderately reduce inference efficiency. Similar trends are also observed on RSVQA-LR and RSIVQA. Combined with the accuracy trends discussed above, these observations suggest a clear trade-off between performance and efficiency: moderate block partitioning improves spatial discrimination, whereas excessively fine partitioning increases latency while providing limited or no accuracy improvement. Accordingly, we adopt $N=4$ for RSVQA-LR and RSIVQA and $N=16$ for RSVQA-HR in the main experiments, as these settings better match the spatial granularity required by different input resolution regimes while maintaining a favorable balance between accuracy and inference cost.

\begin{table}[t]
\scriptsize
\centering
\setlength{\tabcolsep}{3.5pt}
\renewcommand{\arraystretch}{1.3}
\caption{Ablation results comparing original GRASP (baseline) and the uniform soft prompt variant (without question guidance) on Qwen2.5-VL-7B. \textcolor{red}{Red} values indicate performance drops relative to GRASP. \textbf{Bold} marks the best results.}
\label{tab:ablation_uniform}
\begin{tabular}{l|c|c}
\hline
\textbf{Metric} & \textbf{GRASP (Baseline)} & \textbf{Uniform Prompts} \\
\hline
\multicolumn{3}{c}{\textbf{RSVQA-LR}} \\
\hline
\textbf{Count} & \textbf{31.1} & 26.4 \textcolor{red}{(-4.7)} \\
\textbf{R/U} & \textbf{82.6} & 78.3 \textcolor{red}{(-4.3)} \\
\textbf{Pres.} & \textbf{81.6} & 79.0 \textcolor{red}{(-2.6)} \\
\textbf{Comp.} & \textbf{81.5} & 77.9 \textcolor{red}{(-3.6)} \\
\textbf{AA} & \textbf{69.2} & 65.4 \textcolor{red}{(-3.8)} \\
\hline
\multicolumn{3}{c}{\textbf{RSVQA-HR}} \\
\hline
\textbf{Count} & \textbf{62.7} & 58.6 \textcolor{red}{(-4.1)} \\
\textbf{Area} & \textbf{71.0} & 67.6 \textcolor{red}{(-3.4)} \\
\textbf{Pres.} & \textbf{73.9} & 71.7 \textcolor{red}{(-2.2)} \\
\textbf{Comp.} & \textbf{82.5} & 79.9 \textcolor{red}{(-2.6)} \\
\textbf{AA} & \textbf{72.5} & 69.5 \textcolor{red}{(-3.0)} \\
\hline
\multicolumn{3}{c}{\textbf{RSIVQA}} \\
\hline
\textbf{Yes/No} & \textbf{93.1} & 88.7 \textcolor{red}{(-4.4)} \\
\textbf{Num.} & \textbf{61.5} & 56.1 \textcolor{red}{(-5.4)} \\
\textbf{Other} & \textbf{89.0} & 84.7 \textcolor{red}{(-4.3)} \\
\textbf{AA} & \textbf{81.2} & 76.5 \textcolor{red}{(-4.7)} \\
\hline
\end{tabular}
\end{table}

\textbf{Question-guided Sparse Fusion.} Table~\ref{tab:ablation_uniform} shows that replacing the question-guided sparse fusion with uniform prompts (i.e., assigning equal weights to all block-wise prompts when forming $p_{\text{global}}$) consistently reduces accuracy across all datasets, with overall drops of 3.0--4.7 points. The degradation becomes more pronounced on Counting, Area, and Number questions, which demand precise localization and selective aggregation of spatial evidence. Uniform prompting assigns equal contributions to all spatial blocks, so background-dominated regions inject irrelevant cues into the global prompt.
In contrast, GRASP conditions the fusion weights on the input question and sparsifies the aggregation, which suppresses irrelevant regions and concentrates capacity on query-relevant areas. These results indicate that GRASP improves RSVQA not only by introducing block-wise prompts, but by enabling query-dependent spatial selection that supports robust reasoning in cluttered RS scenes.

\textbf{Effect of Entmax Sparsity Parameter.} We analyze the impact of the Entmax sparsity parameter $\alpha$, which governs the density of the attention distribution during prompt fusion. As Table~\ref{tab:ablation_alpha_unified} presents, the model performance exhibits an inverted U-shaped trend as $\alpha$ increases from 1.0 to 2.0 across all three datasets. The standard Softmax function ($\alpha=1.0$) yields the lowest performance (e.g., 65.9\% AA on RSVQA-LR), primarily because it assigns non-zero probabilities to all spatial regions. This forces the model to incorporate irrelevant background noise, which is particularly detrimental in complex RS scenes. Increasing $\alpha$ introduces necessary sparsity, which effectively filters out these distractions. 

Performance peaks at $\alpha=1.5$, where the model achieves the highest accuracy on all benchmarks (69.2\%, 72.5\%, and 81.2\% AA, respectively). This suggests that $\alpha=1.5$ strikes an optimal balance, enabling the model to focus on salient regions while retaining sufficient contextual information. However, further increasing $\alpha$ to 2.0 (Sparsemax) leads to a performance decline. This indicates that excessive sparsity causes the model to discard potentially useful peripheral context, resulting in information loss. Consequently, we select $\alpha=1.5$ as the default setting to ensure robust noise suppression without compromising semantic completeness.

Overall, the ablation results exhibit a consistent structural pattern, indicating that GRASP is primarily driven by its architectural design rather than sensitive hyperparameter tuning. Performance shows a clear drop when the question-guided fusion module is removed, highlighting its role in adaptive spatial selection, while results remain stable across different spatial block granularities and Entmax sparsity settings, suggesting robustness to partition and sparsity configurations.

\subsection{Extended Evaluation}

\textbf{Scene Classification.} We evaluate GRASP on three classic remote sensing scene classification benchmarks, namely EuroSAT~\cite{helber2019eurosat}, AID~\cite{xia2017aid}, and UCMerced~\cite{yang2010bag}. EuroSAT is a satellite image dataset for land-use and land-cover classification, while AID and UCMerced are widely used aerial scene classification benchmarks covering diverse semantic categories. To align with the generative nature of MLLMs, we reformulate scene classification as a visual question answering task, where the model generates the scene category name. We consider two settings: (1) guided, where candidate class names are provided in the prompt, and (2) open-ended, where no candidates are given. For EuroSAT and UCMerced, images are resized to $224 \times 224$ with aspect-ratio-preserving padding, consistent with the low-resolution protocol in our main experiments. For AID, we use a higher input resolution of $512 \times 512$ to better preserve its richer scene structure and spatial details. All images follow aspect-ratio-preserving resizing and center padding to avoid geometric distortion.

Results are presented in Table~\ref{tab:scene_classification_comparison}. GRASP achieves the best performance in the guided setting across all datasets and remains competitive in the open-ended setting, demonstrating its ability to capture scene-level semantics effectively, especially when provided with explicit semantic guidance.
\begin{table}[h]
\centering
\caption{Scene Classification Performance (Accuracy \%) under Guided / Open-ended (G/O) Prompts. Guided: with candidate options; Open-ended: without options. \textbf{Bold} indicates the best performance.}
\label{tab:scene_classification_comparison}
\footnotesize
\resizebox{0.9\columnwidth}{!}{
\begin{tabular}{lccc}
\toprule
\textbf{Method} & \makecell{\textbf{EuroSAT} \\ (G/O)} & \makecell{\textbf{UCMerced} \\ (G/O)} & \makecell{\textbf{AID} \\ (G/O)} \\
\midrule
Qwen2.5-VL (Base) & 48.2 / 19.5 & 65.8 / 14.3 & 58.5 / 13.1 \\
Prompt Tuning     & 55.1 / 22.4 & 66.5 / 18.0 & 60.3 / 19.4 \\
APrompt           & 59.5 / 28.7 & 70.4 / 20.6 & 62.1 / 21.5 \\
VPT               & 58.4 / 30.3 & 69.2 / 26.4 & 65.7 / 32.8 \\
LoRA              & 64.1 / 35.2 & 76.5 / 31.4 & 67.0 / \textbf{38.7} \\
DoRA              & 65.6 / \textbf{36.3} & 77.2 / \textbf{32.0} & 67.4 / 38.2 \\

\textbf{GRASP (ours)} & \textbf{68.1} / 32.6 & \textbf{81.8} / 27.5 & \textbf{68.7} / 33.8 \\
\bottomrule
\end{tabular}
}
\end{table}

\textbf{Visual Grounding.} We further assess GRASP on visual grounding using two remote sensing referring expression benchmarks, VRSBench and DIOR-RSVG. VRSBench provides large-scale remote sensing visual grounding annotations for phrase-to-region alignment, while DIOR-RSVG is built upon DIOR and focuses on referring expression based object localization in complex aerial scenes. The task requires generating normalized bounding boxes in the format $<|box_start|>(x1,y1),(x2,y2)<|box_end|>$ based on referring expressions. All images are resized to $512 \times 512$ with padding for consistent spatial partitioning. We report Acc@0.5, Acc@0.7, and mIoU.
\begin{table}[h]
\centering
\caption{Performance comparison on Visual Grounding task across VRSBench and DIOR-RSVG datasets. \textbf{Bold} indicates the best performance.}
\label{tab:vg_results_compact}
\scriptsize
\setlength{\tabcolsep}{12pt} 
\renewcommand{\arraystretch}{1.1}

\begin{tabular}{l c c c}
\toprule
\textbf{Method} & Acc@0.5 & Acc@0.7 & mIoU \\

\midrule
\rowcolor{gray!15}
\multicolumn{4}{c}{\textbf{VRSBench}} \\ 

Qwen2.5-VL (Base) & 42.5 & 23.9 & 41.7 \\
Prompt Tuning     & 47.3 & 29.8 & 43.6 \\
APrompt           & 50.6 & 30.9 & 46.8 \\
VPT               & 56.2 & 34.5 & 50.9 \\
LoRA              & 60.7 & \underline{38.1} & 53.8 \\
DoRA              & \underline{61.6} & 37.6 & \underline{54.3} \\
\textbf{GRASP (ours)} & \textbf{63.0} & \textbf{40.5} & \textbf{55.6} \\

\rowcolor{gray!15}
\multicolumn{4}{c}{\textbf{DIOR-RSVG}} \\

Qwen2.5-VL (Base) & 40.2 & 26.7 & 39.4 \\
Prompt Tuning     & 46.8 & 31.9 & 43.7 \\
APrompt           & 50.3 & 35.4 & 46.2 \\
VPT               & 56.7 & 40.8 & 51.5 \\
LoRA              & 67.2 & 50.1 & 58.6 \\
DoRA              & \underline{68.4} & \underline{51.3} & \underline{59.4} \\
\textbf{GRASP (ours)} & \textbf{71.6} & \textbf{54.2} & \textbf{62.7} \\
\bottomrule
\end{tabular}
\end{table}

\begin{table}[h]
\centering
\caption{Scalability analysis on Mini-InternVL-4B. \textbf{Bold} indicates the best performance for each dataset. Count and Area tasks are excluded following the official evaluation protocol.}
\label{tab:rsvqa_4b_scalability}
\scriptsize
\setlength{\tabcolsep}{3pt}
\renewcommand{\arraystretch}{1.2}
\begin{tabular}{l | c c c c | c c c}
\toprule
\multirow{2}{*}{\textbf{Model}} & \multicolumn{4}{c|}{\textbf{RSVQA-LR}} & \multicolumn{3}{c}{\textbf{RSVQA-HR}} \\
& \textbf{R/U} & \textbf{Pres.} & \textbf{Comp.} & \textbf{AA (\%)} & \textbf{Pres.} & \textbf{Comp.} & \textbf{AA (\%)} \\
\midrule
Mini-InternVL-4B & 64.7 & 63.1 & 71.8 & 66.5 & 61.0 & 78.0 & 69.5 \\
Prompt Tuning             & 66.2 & 64.5 & 73.0 & 67.9 & 62.5 & 79.4 & 71.0 \\
APrompt                   & 67.5 & 65.8 & 74.2 & 69.2 & 63.8 & 80.7 & 72.3 \\
VPT                       & 69.4 & 66.5 & 75.1 & 70.0 & 65.2 & 81.5 & 73.4 \\
LoRA                      & 73.8 & 72.5 & 82.4 & 76.2 & 75.6 & 88.3 & 82.0 \\
DoRA                      & 74.5 & 73.2 & 83.1 & 76.9 & 76.5 & 89.1 & 82.8 \\
\midrule
\textbf{GRASP (Ours)}     & \textbf{76.8} & \textbf{75.4} & \textbf{86.3} & \textbf{79.5} & \textbf{81.2} & \textbf{91.7} & \textbf{86.5} \\
\bottomrule
\end{tabular}
\end{table}
As shown in Table~\ref{tab:vg_results_compact}, GRASP outperforms baselines on both datasets, indicating strong region-level localization capability transferred from the RSVQA design. These cross-task results confirm that GRASP's spatially explicit prompting provides benefits beyond VQA, particularly in tasks requiring precise region-semantic alignment.

Across these additional evaluations, we observe that the advantage of GRASP is more evident when the task provides explicit semantic guidance, such as candidate labels in guided scene classification. We do not view this sensitivity as a limitation; instead, it supports the effectiveness of the proposed question-guided sparse fusion mechanism. In remote sensing imagery, relevant targets are often sparse and surrounded by large amounts of background content, which makes the visual signal-to-noise ratio of a frozen MLLM relatively low. Under such conditions, clearer semantic anchors in the query help the model assign more discriminative sparse weights over spatial blocks, thereby improving its ability to suppress irrelevant regions and focus on informative local evidence. Overall, these results further highlight the importance of explicitly coupling task semantics with region-structured prompting for remote sensing reasoning.

\textbf{Robustness across datasets and backbones.}
To examine whether the benefit of GRASP extends beyond the two main 7B backbones, we further evaluate it on Mini-InternVL-4B, a lighter model from a different architecture family. As shown in Table~\ref{tab:rsvqa_4b_scalability}, GRASP consistently outperforms the compared PEFT baselines on both RSVQA-LR and RSVQA-HR, achieving the highest overall accuracy in both settings. This gain is partly attributable to the fact that lightweight MLLMs are more susceptible to background noise and limited spatial discrimination in remote sensing imagery, whereas GRASP provides a more explicit mechanism for selecting question-relevant local evidence. These results indicate that the effectiveness of region-aware sparse prompting is not restricted to the original 7B setting and can transfer effectively to more lightweight backbones.

\begin{table}[h]
\centering
\caption{Cross-dataset generalization results. $\Delta$ is defined as cross-dataset accuracy minus in-domain accuracy. Negative values indicate performance degradation after cross-dataset transfer; values closer to zero indicate better robustness.}
\label{tab:cross_dataset}
\scriptsize
\setlength{\tabcolsep}{1.4pt}
\renewcommand{\arraystretch}{1}
\begin{tabular}{lccc|ccc}
\toprule
\multirow{2}{*}{\textbf{Method}} 
& \multicolumn{3}{c|}{\textbf{RSVQA-LR $\rightarrow$ RSIVQA}} 
& \multicolumn{3}{c}{\textbf{RSIVQA $\rightarrow$ RSVQA-LR}} \\
& In-domain & Cross-dataset & $\Delta$
& In-domain & Cross-dataset & $\Delta$ \\
\midrule
Prompt Tuning & 61.1 & 48.3 & -12.8 & 71.5 & 55.8 & -15.7 \\
VPT           & 67.5 & 57.2 & -10.3 & 77.9 & 65.2 & -12.7 \\
LoRA          & 68.7 & 60.9 & -7.8  & 80.7 & 72.1 & -8.6 \\
\textbf{GRASP} & 69.2 & 62.3 & \textbf{-6.9} & 81.2 & 74.4 & \textbf{-6.8} \\
\bottomrule
\end{tabular}
\end{table}

We further evaluate cross-dataset generalization by training each method on one benchmark and directly testing it on another without target-domain fine-tuning. Specifically, we consider two transfer directions, RSVQA-LR $\rightarrow$ RSIVQA and RSIVQA $\rightarrow$ RSVQA-LR.  As shown in Table~\ref{tab:cross_dataset}, all methods suffer a performance drop when transferred across datasets, reflecting the distribution gap between the two benchmarks. However, GRASP consistently shows the smallest degradation in both directions, with $\Delta=-6.9$ for RSVQA-LR $\rightarrow$ RSIVQA and $\Delta=-6.8$ for RSIVQA $\rightarrow$ RSVQA-LR. Compared with Prompt Tuning, VPT, and LoRA, this smaller performance drop suggests that the region-structured prompting and question-guided sparse fusion mechanisms of GRASP provide better resistance to cross-dataset shift. Rather than relying only on globally shared adaptation, GRASP encourages the model to select semantically relevant local evidence, which appears to transfer more robustly across datasets with different image characteristics and question distributions.

\section{Discussion}
\subsection{Fixed Grid vs. Adaptive Partitioning}

To examine whether fixed-grid partitioning is sufficient for heterogeneous remote sensing scenes, we further compare the default GRASP design with an adaptive partitioning strategy. The adaptive variant starts from a coarse grid and recursively refines high-variance regions, while keeping homogeneous areas at a coarser granularity. This is motivated by the fact that RS images often contain both dense urban structures and large low-information background regions.
\begin{table}[h]
\centering
\caption{Systematic analysis of spatial partition configurations on RSVQA-LR and RSVQA-HR datasets. \textbf{Bold} indicates the best performance for each dataset. Latency is reported as the average inference time per sample.}
\label{tab:rsvqa_metric_ablation01}
\scriptsize
\setlength{\tabcolsep}{5pt} 
\renewcommand{\arraystretch}{1.2}

\begin{tabular}{l | c c c c c | c}
\toprule

\rowcolor{gray!15}
\multicolumn{7}{c}{\textbf{RSVQA-LR}} \\

\textbf{Strategy} & \textbf{Count.} & \textbf{R/U} & \textbf{Pres.} & \textbf{Comp.} & \textbf{AA (\%)} & \textbf{Lat. (ms)} \\
\midrule
Fixed Grid  & 31.1 & 82.6 & 81.6 & 81.5 & 69.2 & \textbf{112} \\

Adaptive  & \textbf{32.0} & \textbf{83.0} & \textbf{81.9} & \textbf{82.1} & \textbf{69.7} & 188 \\

\rowcolor{gray!15}
\multicolumn{7}{c}{\textbf{RSVQA-HR}} \\

\textbf{Strategy} & \textbf{Count.} & \textbf{Area} & \textbf{Pres.} & \textbf{Comp.} & \textbf{AA (\%)} & \textbf{Lat. (ms)} \\
\midrule
Fixed Grid & 62.7 & 71.0 & 73.9 & 82.5 & 72.5 & \textbf{194} \\

Adaptive  & \textbf{63.9} & \textbf{72.4} & \textbf{75.5} & \textbf{83.9} & \textbf{73.9} & 274 \\
\bottomrule
\end{tabular}
\end{table}
As shown in Table~\ref{tab:rsvqa_metric_ablation01}, adaptive partitioning improves performance on both RSVQA-LR and RSVQA-HR, with a larger gain on the high-resolution setting. In particular, the improvement is more evident on Count and Area questions, suggesting that adaptive partitioning better preserves fine-grained local structures and small-object details in complex scenes. However, this benefit comes with higher inference cost. Compared with the fixed-grid version, the adaptive strategy increases latency from 112 ms to 188 ms on RSVQA-LR and from 194 ms to 274 ms on RSVQA-HR. Therefore, while adaptive partitioning is a promising extension, the original fixed-grid GRASP still offers a more practical efficiency-effectiveness trade-off under the current setting.

\section{Conclusion}
This paper demonstrates that explicitly modeling spatial structure plays a critical role in adapting general-purpose multimodal large language models to the RS domain. Our observations indicate that adaptation performance does not solely depend on parameter scale or tuning strategies, but is largely influenced by how visual information is spatially organized and queried. The performance of GRASP suggests that lightweight prompting mechanisms with spatial awareness can effectively compensate for the limitations of global adaptation strategies in complex scenes, providing a viable path toward parameter-efficient model transfer. 
Looking ahead, future work may explore more flexible region modeling to address the limitations of fixed grid partitioning. It is also promising to extend the proposed structure-aware prompting paradigm to multi-granularity hierarchical architectures, enabling better scalability to ultra-high-definition remote sensing imagery while maintaining computational efficiency.

\section{Acknowledgement}
This work was supported by the National Natural Science Foundation of China (Grant No. 62572104).

\bibliographystyle{IEEEtran}
\bibliography{refs}

@article{zhang2024jointly,
  title={Jointly RS Image Deblurring and Super-Resolution With Adjustable-Kernel and Multi-Domain Attention},
  author={Zhang, Yan and Zheng, Pengcheng and Zeng, Chengxiao and Xiao, Bin and Li, Zhenghao and Gao, Xinbo},
  journal={IEEE Transactions on Geoscience and Remote Sensing},
  year={2024},
  publisher={IEEE}
}

@article{zheng2023cgc,
  title={CGC-net: A context-guided constrained network for remote-sensing image super resolution},
  author={Zheng, Pengcheng and Jiang, Jianan and Zhang, Yan and Zeng, Chengxiao and Qin, Chuanchuan and Li, Zhenghao},
  journal={Remote Sensing},
  volume={15},
  number={12},
  pages={3171},
  year={2023},
  publisher={MDPI}
}

@article{zhu2017deep,
  title={Deep learning in remote sensing: A comprehensive review and list of resources},
  author={Zhu, Xiao Xiang and Tuia, Devis and Mou, Lichao and Xia, Gui-Song and Zhang, Liangpei and Xu, Feng and Fraundorfer, Friedrich},
  journal={IEEE geoscience and remote sensing magazine},
  volume={5},
  number={4},
  pages={8--36},
  year={2017},
  publisher={IEEE}
}

@incollection{lobry2024visual,
  title={Visual question answering on remote sensing images},
  author={Lobry, Sylvain and Tuia, Devis},
  booktitle={Advances in Machine Learning and Image Analysis for GeoAI},
  pages={237--254},
  year={2024},
  publisher={Elsevier}
}

@article{gomez2015multimodal,
  title={Multimodal classification of remote sensing images: A review and future directions},
  author={G{\'o}mez-Chova, Luis and Tuia, Devis and Moser, Gabriele and Camps-Valls, Gustau},
  journal={Proceedings of the IEEE},
  volume={103},
  number={9},
  pages={1560--1584},
  year={2015},
  publisher={IEEE}
}

@article{li2024vision,
  title={Vision-language models in remote sensing: Current progress and future trends},
  author={Li, Xiang and Wen, Congcong and Hu, Yuan and Yuan, Zhenghang and Zhu, Xiao Xiang},
  journal={IEEE Geoscience and Remote Sensing Magazine},
  year={2024},
  publisher={IEEE}
}

@article{zhou2024towards,
  title={Towards vision-language geo-foundation model: A survey},
  author={Zhou, Yue and Feng, Litong and Ke, Yiping and Jiang, Xue and Yan, Junchi and Yang, Xue and Zhang, Wayne},
  journal={arXiv preprint arXiv:2406.09385},
  year={2024}
}

@article{liu2024remoteclip,
  title={Remoteclip: A vision language foundation model for remote sensing},
  author={Liu, Fan and Chen, Delong and Guan, Zhangqingyun and Zhou, Xiaocong and Zhu, Jiale and Ye, Qiaolin and Fu, Liyong and Zhou, Jun},
  journal={IEEE Transactions on Geoscience and Remote Sensing},
  year={2024},
  publisher={IEEE}
}

@article{lobry2020rsvqa,
  title={RSVQA: Visual question answering for remote sensing data},
  author={Lobry, Sylvain and Marcos, Diego and Murray, Jesse and Tuia, Devis},
  journal={IEEE Transactions on Geoscience and Remote Sensing},
  volume={58},
  number={12},
  pages={8555--8566},
  year={2020},
  publisher={IEEE}
}

@article{lester2021power,
  title={The power of scale for parameter-efficient prompt tuning},
  author={Lester, Brian and Al-Rfou, Rami and Constant, Noah},
  journal={arXiv preprint arXiv:2104.08691},
  year={2021}
}

@inproceedings{siebert2022multi,
  title={Multi-modal fusion transformer for visual question answering in remote sensing},
  author={Siebert, Tim and Clasen, Kai Norman and Ravanbakhsh, Mahdyar and Demir, Beg{\"u}m},
  booktitle={Image and Signal Processing for Remote Sensing XXVIII},
  volume={12267},
  pages={162--170},
  year={2022},
  organization={SPIE}
}

@article{feng2024multi,
  title={A multi-scale contextual attention network for remote sensing visual question answering},
  author={Feng, Jiangfan and Wang, Hui},
  journal={International Journal of Applied Earth Observation and Geoinformation},
  volume={126},
  pages={103641},
  year={2024},
  publisher={Elsevier}
}

@article{lu2019vilbert,
  title={Vilbert: Pretraining task-agnostic visiolinguistic representations for vision-and-language tasks},
  author={Lu, Jiasen and Batra, Dhruv and Parikh, Devi and Lee, Stefan},
  journal={Advances in neural information processing systems},
  volume={32},
  year={2019}
}

@inproceedings{silva2022remote,
  title={Remote sensing visual question answering with a self-attention multi-modal encoder},
  author={Silva, Jo{\~a}o Daniel and Magalh{\~a}es, Jo{\~a}o and Tuia, Devis and Martins, Bruno},
  booktitle={Proceedings of the 5th ACM SIGSPATIAL International Workshop on AI for Geographic Knowledge Discovery},
  pages={40--49},
  year={2022}
}

@article{zhu2024mvp,
  title={Mvp: Meta visual prompt tuning for few-shot remote sensing image scene classification},
  author={Zhu, Junjie and Li, Yiying and Yang, Ke and Guan, Naiyang and Fan, Zunlin and Qiu, Chunping and Yi, Xiaodong},
  journal={IEEE Transactions on Geoscience and Remote Sensing},
  volume={62},
  pages={1--13},
  year={2024},
  publisher={IEEE}
}

@inproceedings{bhattacharya2023c,
  title={C-SAW: self-supervised prompt learning for image generalization in remote sensing},
  author={Bhattacharya, Avigyan and Singha, Mainak and Jha, Ankit and Banerjee, Biplab},
  booktitle={Proceedings of the Fourteenth Indian Conference on Computer Vision, Graphics and Image Processing},
  pages={1--10},
  year={2023}
}

@inproceedings{xu2015show,
  title={Show, attend and tell: Neural image caption generation with visual attention},
  author={Xu, Kelvin and Ba, Jimmy and Kiros, Ryan and Cho, Kyunghyun and Courville, Aaron and Salakhudinov, Ruslan and Zemel, Rich and Bengio, Yoshua},
  booktitle={International conference on machine learning},
  pages={2048--2057},
  year={2015},
  organization={PMLR}
}

@inproceedings{antol2015vqa,
  title={Vqa: Visual question answering},
  author={Antol, Stanislaw and Agrawal, Aishwarya and Lu, Jiasen and Mitchell, Margaret and Batra, Dhruv and Zitnick, C Lawrence and Parikh, Devi},
  booktitle={Proceedings of the IEEE international conference on computer vision},
  pages={2425--2433},
  year={2015}
}

@article{zheng2021mutual,
  title={Mutual attention inception network for remote sensing visual question answering},
  author={Zheng, Xiangtao and Wang, Binqiang and Du, Xingqian and Lu, Xiaoqiang},
  journal={IEEE Transactions on Geoscience and Remote Sensing},
  volume={60},
  pages={1--14},
  year={2021},
  publisher={IEEE}
}

@inproceedings{wang2024earthvqa,
  title={Earthvqa: Towards queryable earth via relational reasoning-based remote sensing visual question answering},
  author={Wang, Junjue and Zheng, Zhuo and Chen, Zihang and Ma, Ailong and Zhong, Yanfei},
  booktitle={Proceedings of the AAAI Conference on Artificial Intelligence},
  volume={38},
  number={6},
  pages={5481--5489},
  year={2024}
}

@inproceedings{chappuis2022prompt,
  title={Prompt-RSVQA: Prompting visual context to a language model for remote sensing visual question answering},
  author={Chappuis, Christel and Zermatten, Val{\'e}rie and Lobry, Sylvain and Le Saux, Bertrand and Tuia, Devis},
  booktitle={Proceedings of the IEEE/CVF conference on computer vision and pattern recognition},
  pages={1372--1381},
  year={2022}
}

@article{yuan2022exploring,
  title={Exploring a fine-grained multiscale method for cross-modal remote sensing image retrieval},
  author={Yuan, Zhiqiang and Zhang, Wenkai and Fu, Kun and Li, Xuan and Deng, Chubo and Wang, Hongqi and Sun, Xian},
  journal={arXiv preprint arXiv:2204.09868},
  year={2022}
}

@inproceedings{zhou2022conditional,
  title={Conditional prompt learning for vision-language models},
  author={{Zhou, Kaiyang and Yang, Jingkang and Loy, Chen Change and Liu, Ziwei}},
  booktitle={Proceedings of the IEEE/CVF conference on computer vision and pattern recognition},
  pages={16816--16825},
  year={2022}
}

@inproceedings{jia2022visual,
  title={Visual prompt tuning},
  author={Jia, Menglin and Tang, Luming and Chen, Bor-Chun and Cardie, Claire and Belongie, Serge and Hariharan, Bharath and Lim, Ser-Nam},
  booktitle={European conference on computer vision},
  pages={709--727},
  year={2022},
  organization={Springer}
}

@misc{liu2024llavanext,
    title={LLaVA-NeXT: Improved reasoning, OCR, and world knowledge},
    url={https://llava-vl.github.io/blog/2024-01-30-llava-next/},
    author={Liu, Haotian and Li, Chunyuan and Li, Yuheng and Li, Bo and Zhang, Yuanhan and Shen, Sheng and Lee, Yong Jae},
    month={January},
    year={2024}
}

@article{bai2025qwen2,
  title={Qwen2. 5-vl technical report},
  author={Bai, Shuai and Chen, Keqin and Liu, Xuejing and Wang, Jialin and Ge, Wenbin and Song, Sibo and Dang, Kai and Wang, Peng and Wang, Shijie and Tang, Jun and others},
  journal={arXiv preprint arXiv:2502.13923},
  year={2025}
}

@inproceedings{kuckreja2024geochat,
  title={Geochat: Grounded large vision-language model for remote sensing},
  author={Kuckreja, Kartik and Danish, Muhammad Sohail and Naseer, Muzammal and Das, Abhijit and Khan, Salman and Khan, Fahad Shahbaz},
  booktitle={Proceedings of the IEEE/CVF Conference on Computer Vision and Pattern Recognition},
  pages={27831--27840},
  year={2024}
}

@ARTICLE{zhan2025skyeyegpt,
      title={SkyEyeGPT: Unifying Remote Sensing Vision-Language Tasks via Instruction Tuning with Large Language Model}, 
      author={Yang Zhan and Zhitong Xiong and Yuan Yuan},
      year={2025},
      journal={ISPRS Journal of Photogrammetry and Remote Sensing},
      volume = {221},
      pages = {64-77}
}

@article{bazi2024rs,
  title={Rs-llava: A large vision-language model for joint captioning and question answering in remote sensing imagery},
  author={Bazi, Yakoub and Bashmal, Laila and Al Rahhal, Mohamad Mahmoud and Ricci, Riccardo and Melgani, Farid},
  journal={Remote Sensing},
  volume={16},
  number={9},
  pages={1477},
  year={2024},
  publisher={MDPI}
}

@article{yin2024survey,
  title={A survey on multimodal large language models},
  author={Yin, Shukang and Fu, Chaoyou and Zhao, Sirui and Li, Ke and Sun, Xing and Xu, Tong and Chen, Enhong},
  journal={National Science Review},
  volume={11},
  number={12},
  pages={nwae403},
  year={2024},
  publisher={Oxford University Press}
}

@article{zhang2024mm,
  title={Mm-llms: Recent advances in multimodal large language models},
  author={Zhang, Duzhen and Yu, Yahan and Dong, Jiahua and Li, Chenxing and Su, Dan and Chu, Chenhui and Yu, Dong},
  journal={arXiv preprint arXiv:2401.13601},
  year={2024}
}

@article{hu2022lora,
  title={Lora: Low-rank adaptation of large language models.},
  author={Hu, Edward J and Shen, Yelong and Wallis, Phillip and Allen-Zhu, Zeyuan and Li, Yuanzhi and Wang, Shean and Wang, Lu and Chen, Weizhu and others},
  journal={ICLR},
  volume={1},
  number={2},
  pages={3},
  year={2022}
}

@inproceedings{wang2023aprompt,
  title={Aprompt: Attention prompt tuning for efficient adaptation of pre-trained language models},
  author={Wang, Qifan and Mao, Yuning and Wang, Jingang and Yu, Hanchao and Nie, Shaoliang and Wang, Sinong and Feng, Fuli and Huang, Lifu and Quan, Xiaojun and Xu, Zenglin and others},
  booktitle={Proceedings of the 2023 conference on empirical methods in natural language processing},
  pages={9147--9160},
  year={2023}
}

@inproceedings{houlsby2019parameter,
  title={Parameter-efficient transfer learning for NLP},
  author={Houlsby, Neil and Giurgiu, Andrei and Jastrzebski, Stanislaw and Morrone, Bruna and De Laroussilhe, Quentin and Gesmundo, Andrea and Attariyan, Mona and Gelly, Sylvain},
  booktitle={International conference on machine learning},
  pages={2790--2799},
  year={2019},
  organization={PMLR}
}

@inproceedings{xia2018dota,
  title={DOTA: A large-scale dataset for object detection in aerial images},
  author={Xia, Gui-Song and Bai, Xiang and Ding, Jian and Zhu, Zhen and Belongie, Serge and Luo, Jiebo and Datcu, Mihai and Pelillo, Marcello and Zhang, Liangpei},
  booktitle={Proceedings of the IEEE conference on computer vision and pattern recognition},
  pages={3974--3983},
  year={2018}
}

@article{li2025ddfav,
  title={Ddfav: Remote sensing large vision language models dataset and evaluation benchmark},
  author={Li, Haodong and Zhang, Xiaofeng and Qu, Haicheng},
  journal={Remote Sensing},
  volume={17},
  number={4},
  pages={719},
  year={2025},
  publisher={MDPI}
}

@article{zhang2024earthgpt,
  title={EarthGPT: A universal multimodal large language model for multisensor image comprehension in remote sensing domain},
  author={Zhang, Wei and Cai, Miaoxin and Zhang, Tong and Zhuang, Yin and Mao, Xuerui},
  journal={IEEE Transactions on Geoscience and Remote Sensing},
  volume={62},
  pages={1--20},
  year={2024},
  publisher={IEEE}
}

@article{ou2025geopix,
  title={GeoPix: A multimodal large language model for pixel-level image understanding in remote sensing},
  author={Ou, Ruizhe and Hu, Yuan and Zhang, Fan and Chen, Jiaxin and Liu, Yu},
  journal={IEEE Geoscience and Remote Sensing Magazine},
  year={2025},
  publisher={IEEE}
}

@article{huang2025survey,
  title={A survey on remote sensing foundation models: From vision to multimodality},
  author={Huang, Ziyue and Yan, Hongxi and Zhan, Qiqi and Yang, Shuai and Zhang, Mingming and Zhang, Chenkai and Lei, YiMing and Liu, Zeming and Liu, Qingjie and Wang, Yunhong},
  journal={arXiv preprint arXiv:2503.22081},
  year={2025}
}

@inproceedings{liu2024improved,
  title={Improved baselines with visual instruction tuning},
  author={Liu, Haotian and Li, Chunyuan and Li, Yuheng and Lee, Yong Jae},
  booktitle={Proceedings of the IEEE/CVF conference on computer vision and pattern recognition},
  pages={26296--26306},
  year={2024}
}

@article{sun2024pixels,
  title={From Pixels to Prose: Advancing Multi-Modal Language Models for Remote Sensing},
  author={Sun, Xintian and Peng, Benji and Zhang, Charles and Jin, Fei and Niu, Qian and Liu, Junyu and Chen, Keyu and Li, Ming and Feng, Pohsun and Bi, Ziqian and others},
  journal={arXiv preprint arXiv:2411.05826},
  year={2024}
}

@article{zhang2025earthgpt,
  title={EarthGPT-X: A Spatial MLLM for Multi-level Multi-Source Remote Sensing Imagery Understanding with Visual Prompting},
  author={Zhang, Wei and Cai, Miaoxin and Ning, Yaqian and Zhang, Tong and Zhuang, Yin and Lu, Shijian and Chen, He and Li, Jun and Mao, Xuerui},
  journal={IEEE Transactions on Geoscience and Remote Sensing},
  year={2025},
  publisher={Institute of Electrical and Electronics Engineers Inc.}
}

@article{peters2019sparse,
  title={Sparse sequence-to-sequence models},
  author={Peters, Ben and Niculae, Vlad and Martins, Andr{\'e} FT},
  journal={arXiv preprint arXiv:1905.05702},
  year={2019}
}

@article{pang2024h2rsvlm,
  title={H2rsvlm: Towards helpful and honest remote sensing large vision language model},
  author={Pang, Chao and Wu, Jiang and Li, Jiayu and Liu, Yi and Sun, Jiaxing and Li, Weijia and Weng, Xingxing and Wang, Shuai and Feng, Litong and Xia, Gui-Song and others},
  journal={CoRR},
  year={2024}
}

@inproceedings{hackel2023lit,
  title={Lit-4-rsvqa: Lightweight transformer-based visual question answering in remote sensing},
  author={Hackel, Leonard and Clasen, Kai Norman and Ravanbakhsh, Mahdyar and Demir, Beg{\"u}m},
  booktitle={IGARSS 2023-2023 IEEE International Geoscience and Remote Sensing Symposium},
  pages={2231--2234},
  year={2023},
  organization={IEEE}
}

@inproceedings{liu2024dora,
  title={Dora: Weight-decomposed low-rank adaptation},
  author={Liu, Shih-Yang and Wang, Chien-Yi and Yin, Hongxu and Molchanov, Pavlo and Wang, Yu-Chiang Frank and Cheng, Kwang-Ting and Chen, Min-Hung},
  booktitle={Forty-first International Conference on Machine Learning},
  year={2024}
}

@inproceedings{he2016deep,
  title={Deep residual learning for image recognition},
  author={He, Kaiming and Zhang, Xiangyu and Ren, Shaoqing and Sun, Jian},
  booktitle={Proceedings of the IEEE conference on computer vision and pattern recognition},
  pages={770--778},
  year={2016}
}

@article{zhang2024earthmarker,
  title={Earthmarker: A visual prompting multi-modal large language model for remote sensing},
  author={Zhang, Wei and Cai, Miaoxin and Zhang, Tong and Zhuang, Yin and Li, Jun and Mao, Xuerui},
  journal={IEEE Transactions on Geoscience and Remote Sensing},
  year={2024},
  publisher={IEEE}
}

@article{siripong2024large,
  title={Large vision-language models for remote sensing visual question answering},
  author={Siripong, Surasakdi and Chaiyapan, Apirak and Phonchai, Thanakorn},
  journal={arXiv preprint arXiv:2411.10857},
  year={2024}
}

@inproceedings{songara2023visual,
  title={Visual question answering in remote sensing with cross-attention and multimodal information bottleneck},
  author={Songara, Jayesh and Pande, Shivam and Choudhury, Shabnam and Banerjee, Biplab and Velmurugan, Rajbabu},
  booktitle={IGARSS 2023-2023 IEEE International Geoscience and Remote Sensing Symposium},
  pages={6278--6281},
  year={2023},
  organization={IEEE}
}

@article{xiao2025foundation,
  title={Foundation models for remote sensing and earth observation: A survey},
  author={Xiao, Aoran and Xuan, Weihao and Wang, Junjue and Huang, Jiaxing and Tao, Dacheng and Lu, Shijian and Yokoya, Naoto},
  journal={IEEE Geoscience and Remote Sensing Magazine},
  year={2025},
  publisher={IEEE}
}

@article{gao2024mini,
  title={Mini-internvl: a flexible-transfer pocket multi-modal model with 5\% parameters and 90\% performance},
  author={Gao, Zhangwei and Chen, Zhe and Cui, Erfei and Ren, Yiming and Wang, Weiyun and Zhu, Jinguo and Tian, Hao and Ye, Shenglong and He, Junjun and Zhu, Xizhou and others},
  journal={Visual Intelligence},
  volume={2},
  number={1},
  pages={32},
  year={2024},
  publisher={Springer}
}

@article{han2024parameter,
  title={Parameter-efficient fine-tuning for large models: A comprehensive survey},
  author={Han, Zeyu and Gao, Chao and Liu, Jinyang and Zhang, Jeff and Zhang, Sai Qian},
  journal={arXiv preprint arXiv:2403.14608},
  year={2024}
}

@article{ding2023parameter,
  title={Parameter-efficient fine-tuning of large-scale pre-trained language models},
  author={Ding, Ning and Qin, Yujia and Yang, Guang and Wei, Fuchao and Yang, Zonghan and Su, Yusheng and Hu, Shengding and Chen, Yulin and Chan, Chi-Min and Chen, Weize and others},
  journal={Nature machine intelligence},
  volume={5},
  number={3},
  pages={220--235},
  year={2023},
  publisher={Nature Publishing Group UK London}
}

@article{wang2025parameter,
  title={Parameter-efficient fine-tuning in large language models: a survey of methodologies},
  author={Wang, Luping and Chen, Sheng and Jiang, Linnan and Pan, Shu and Cai, Runze and Yang, Sen and Yang, Fei},
  journal={Artificial Intelligence Review},
  volume={58},
  number={8},
  pages={227},
  year={2025},
  publisher={Springer}
}

@article{zhou2025advances,
  title={Advances on multimodal remote sensing foundation models for Earth observation downstream tasks: A survey},
  author={Zhou, Guoqing and Lihuang, Qian and Gamba, Paolo},
  journal={Remote Sensing},
  volume={17},
  number={21},
  pages={3532},
  year={2025},
  publisher={MDPI AG}
}

@article{zhai2023investigating,
  title={Investigating the catastrophic forgetting in multimodal large language models},
  author={Zhai, Yuexiang and Tong, Shengbang and Li, Xiao and Cai, Mu and Qu, Qing and Lee, Yong Jae and Ma, Yi},
  journal={arXiv preprint arXiv:2309.10313},
  year={2023}
}

@inproceedings{tian2025meta,
  title={Meta-Learning Hyperparameters for Parameter Efficient Fine-Tuning},
  author={Tian, Zichen and Liu, Yaoyao and Sun, Qianru},
  booktitle={Proceedings of the Computer Vision and Pattern Recognition Conference},
  pages={23037--23047},
  year={2025}
}

@article{chen2025dynamicvis,
  title={Dynamicvis: An efficient and general visual foundation model for remote sensing image understanding},
  author={Chen, Keyan and Liu, Chenyang and Chen, Bowen and Li, Wenyuan and Zou, Zhengxia and Shi, Zhenwei},
  journal={arXiv preprint arXiv:2503.16426},
  year={2025}
}

@article{zidi2025lola,
  title={LoLA-SpecViT: Local attention SwiGLU vision transformer with LoRA for hyperspectral imaging},
  author={Zidi, Fadi Abdeladhim and Boukhari, Djamel Eddine and Sellam, Abdellah Zakaria and Ouafi, Abdelkrim and Distante, Cosimo and Bekhouche, Salah Eddine and Taleb-Ahmed, Abdelmalik},
  journal={International Journal of Applied Earth Observation and Geoinformation},
  volume={144},
  pages={104924},
  year={2025},
  publisher={Elsevier}
}

@article{xiao2025prompt,
  title={Prompt-based adaptation in large-scale vision models: A survey},
  author={Xiao, Xi and Zhang, Yunbei and Zhao, Lin and Liu, Yiyang and Liao, Xiaoying and Mai, Zheda and Li, Xingjian and Wang, Xiao and Xu, Hao and Hamm, Jihun and others},
  journal={arXiv preprint arXiv:2510.13219},
  year={2025}
}

@article{zhao2025text,
  title={Text-Guided Coarse-to-Fine Fusion Network for robust remote sensing visual question answering},
  author={Zhao, Zhicheng and Zhou, Changfu and Zhang, Yu and Li, Chenglong and Ma, Xiaoliang and Tang, Jin},
  journal={ISPRS Journal of Photogrammetry and Remote Sensing},
  volume={230},
  pages={1--17},
  year={2025},
  publisher={Elsevier}
}

@article{bi2025ringmoe,
  title={RingMoE: Mixture-of-modality-experts multi-modal foundation models for universal remote sensing image interpretation},
  author={Bi, Hanbo and Feng, Yingchao and Tong, Boyuan and Wang, Mengyu and Yu, Haichen and Mao, Yongqiang and Chang, Hao and Diao, Wenhui and Wang, Peijin and Yu, Yue and others},
  journal={IEEE Transactions on Pattern Analysis and Machine Intelligence},
  year={2025},
  publisher={IEEE}
}

@article{helber2019eurosat,
  title={Eurosat: A novel dataset and deep learning benchmark for land use and land cover classification},
  author={Helber, Patrick and Bischke, Benjamin and Dengel, Andreas and Borth, Damian},
  journal={IEEE Journal of Selected Topics in Applied Earth Observations and Remote Sensing},
  volume={12},
  number={7},
  pages={2217--2226},
  year={2019},
  publisher={IEEE}
}

@article{xia2017aid,
  title={AID: A benchmark data set for performance evaluation of aerial scene classification},
  author={Xia, Gui-Song and Hu, Jingwen and Hu, Fan and Shi, Baoguang and Bai, Xiang and Zhong, Yanfei and Zhang, Liangpei and Lu, Xiaoqiang},
  journal={IEEE Transactions on Geoscience and Remote Sensing},
  volume={55},
  number={7},
  pages={3965--3981},
  year={2017},
  publisher={IEEE}
}

@inproceedings{yang2010bag,
  title={Bag-of-visual-words and spatial extensions for land-use classification},
  author={Yang, Yi and Newsam, Shawn},
  booktitle={Proceedings of the 18th SIGSPATIAL international conference on advances in geographic information systems},
  pages={270--279},
  year={2010}
}

@inproceedings{zi2025rsvlm,
  title={RSVLM-QA: A Benchmark Dataset for Remote Sensing Vision Language Model-based Question Answering},
  author={Zi, Xing and Xiao, Jinghao and Shi, Yunxiao and Tao, Xian and Li, Jun and Braytee, Ali and Prasad, Mukesh},
  booktitle={Proceedings of the 33rd ACM International Conference on Multimedia},
  pages={12905--12911},
  year={2025}
}

\newpage

\vfill

\end{document}